\documentclass[final]{cvpr}

\usepackage{times}
\usepackage{epsfig}
\usepackage{graphicx}
\usepackage{amsmath}
\usepackage{amssymb}

\usepackage[square,numbers,sort&compress]{natbib}
\usepackage[pagebackref=true,breaklinks=true,letterpaper=true,colorlinks,bookmarks=false,citecolor=blue]{hyperref}
\usepackage{times,graphicx,tabulary,multirow,xspace}

\usepackage{amssymb}
\usepackage{amsmath,amsfonts}
\usepackage{amsopn,bm}

\usepackage{nicefrac}       

\usepackage{epsfig}
\usepackage{graphicx}
\usepackage{float}
\usepackage{wrapfig}
\usepackage[ruled,vlined]{algorithm2e}

\usepackage{bm,xspace}
\usepackage{comment}
\usepackage{verbatim}
\usepackage{multirow}
\usepackage{makecell}
\usepackage{array}
\usepackage{balance}
\usepackage{url}
\usepackage{booktabs}
\usepackage{etoolbox,siunitx}
\usepackage{calc}
\usepackage{pifont,hologo}
 
\usepackage{nicefrac}

\usepackage{arydshln}
\usepackage[utf8]{inputenc} 
\usepackage[T1]{fontenc}    
\usepackage{url}            
\usepackage{amsfonts}       
\usepackage{nicefrac}       
\usepackage{microtype}      
\usepackage[american]{babel}
\usepackage{enumitem}
\usepackage{siunitx}

\usepackage{enumitem}

\usepackage[super]{nth}
\usepackage{nicefrac}
\robustify\bfseries

\usepackage{dsfont}

\usepackage{listings}
\usepackage{xcolor}
\definecolor{codegreen}{rgb}{0,0.6,0}
\definecolor{codegray}{rgb}{0.5,0.5,0.5}
\definecolor{codepurple}{rgb}{0.58,0,0.82}
\definecolor{backcolour}{rgb}{1.0,1.0,1.0}

\lstdefinestyle{mystyle}{
    commentstyle=\color{codegreen},
    keywordstyle=\color{magenta},
    numberstyle=\tiny\color{codegray},
    stringstyle=\color{codepurple},
    basicstyle=\ttfamily\footnotesize,
    breakatwhitespace=false,         
    breaklines=true,                 
    captionpos=b,                    
    keepspaces=true,                 
    numbersep=0pt,                  
    showspaces=false,                
    showstringspaces=false,
    showtabs=false,                  
    tabsize=2,
    linewidth=.99\textwidth,
    xleftmargin=0.01cm
}
\usepackage{mathtools}

\newcommand{\cmark}{\text{\ding{51}}}
\newcommand{\xmark}{\text{\ding{55}}}

\newcommand{\vct}[1]{\boldsymbol{#1}} 
\newcommand{\cst}[1]{\mathsf{#1}}  

\newcommand{\norm}[1]{\|#1\|}

\newcommand{\ProbOpr}[1]{\mathbb{#1}}

\newcommand{\expect}[2]{%
\ifthenelse{\equal{#2}{}}{\ProbOpr{E}_{#1}}
{\ifthenelse{\equal{#1}{}}{\ProbOpr{E}\left[#2\right]}{\ProbOpr{E}_{#1}\left[#2\right]}}} 
\newcommand{\var}[2]{%
\ifthenelse{\equal{#2}{}}{\ProbOpr{VAR}_{#1}}
{\ifthenelse{\equal{#1}{}}{\ProbOpr{VAR}\left[#2\right]}{\ProbOpr{VAR}_{#1}\left[#2\right]}}} 

\newcommand{\vh}{\vct{h}}

\newcommand{\vp}{\vct{p}}

\newcommand{\vcts}{\vct{s}}
\newcommand{\vt}{\vct{t}}
\newcommand{\vu}{\vct{u}}
\newcommand{\vv}{\vct{v}}

\newcommand{\vx}{{\vct{x}}}

\newcommand{\cK}{\cst{K}}

\newcommand{\cM}{\cst{M}}
\newcommand{\cN}{\cst{N}}

\newcommand{\vphi}{\vct{\phi}}
\newcommand{\vpsi}{\vct{\psi}}
\newcommand{\vtheta}{\vct{\theta}}

\def\bst{\boldsymbol{t}}

\def\bsx{\boldsymbol{x}}

\def\calD{\mathcal{D}}

\makeatletter
\DeclareRobustCommand\onedot{\futurelet\@let@token\@onedot}
\def\@onedot{\ifx\@let@token.\else.\null\fi\xspace}

\def\eg{\emph{e.g}\onedot} 
\def\ie{\emph{i.e}\onedot} 
 
\def\etc{\emph{etc}\onedot} 
 
\def\etal{\emph{et al}\onedot}
\makeatother

\newcommand{\symtext}[2]{\textsc{#1}#2}
\newcommand{\eat}[1]{{}}

\newcommand\mypara[1]{\vspace{1mm}\noindent\textbf{#1}}

\usepackage[caption=false,font=normalsize]{subfig}
\usepackage{tabu}

\definecolor{Gray}{gray}{0.5}

\newlength\savewidth

\makeatletter\renewcommand\paragraph{\@startsection{paragraph}{4}{\z@}
  {.5em \@plus1ex \@minus.2ex}{-.5em}{\normalfont\normalsize\bfseries}}\makeatother

\newcolumntype{x}[1]{>{\centering\arraybackslash}p{#1pt}}
\newcolumntype{y}[1]{>{\raggedright\arraybackslash}p{#1pt}}
\newcolumntype{z}[1]{>{\raggedleft\arraybackslash}p{#1pt}}

\definecolor{Highlight}{HTML}{39b54a}  

\newcommand{\ourtitle}{{Learning the Best Pooling Strategy for Visual Semantic Embedding}}
\newcommand{\vse}{{{VSE}}\xspace}
\newcommand{\vseinf}{{{VSE}$\infty$}\xspace}
\newcommand{\ourmethod}{\vseinf}
\newcommand{\first}[1]{\textcolor{red}{\bf#1}}
\newcommand{\second}[1]{\textcolor{black}{\bf#1}}
	
\newcommand{\SM}{{Appendix}\xspace}

\graphicspath{{./}}

\def\cvprPaperID{524} 
\def\confYear{CVPR 2021}

\begin{document}

\title{\ourtitle}

\author{Jiacheng Chen$^{1}$\thanks{Authors contributed equally} \qquad Hexiang Hu$^2$\footnotemark[1] \qquad Hao Wu$^1$ \qquad Yuning Jiang$^3$ \qquad Changhu Wang$^1$ \\[6pt]
$^1$ByteDance AI Lab \qquad $^2$University of Southern California \qquad $^3$Alibaba Inc \\}

\maketitle

\begin{abstract}
    Visual Semantic Embedding (VSE) is a dominant approach for vision-language retrieval, which aims at learning a deep embedding space such that visual data are embedded close to their semantic text labels or descriptions. Recent VSE models use complex methods to better contextualize and aggregate multi-modal features into holistic embeddings. However, we discover that surprisingly simple (but carefully selected) global pooling functions (\eg, max pooling) outperform those complex models, across different feature extractors. Despite its simplicity and effectiveness, seeking the best pooling function for different data modality and feature extractor is costly and tedious, especially when the size of features varies (\eg, text, video). Therefore, we propose a Generalized Pooling Operator (GPO), which learns to automatically adapt itself to the best pooling strategy for different features, requiring no manual tuning while staying effective and efficient. We extend the VSE model using this proposed GPO and denote it as {\it \ourmethod}.

    Without bells and whistles, {\it \ourmethod} outperforms previous VSE methods significantly on image-text retrieval benchmarks across popular feature extractors. With a simple adaptation, variants of {\it \ourmethod} further demonstrate its strength by achieving the new state of the art on two video-text retrieval datasets. Comprehensive experiments and visualizations confirm that GPO always discovers the best pooling strategy and can be a plug-and-play feature aggregation module for standard VSE models. Code and pre-trained models are available at \url{https://vse-infty.github.io}
\end{abstract}

\section{Introduction}
\label{sec:intro}

Recognizing and describing the visual world with natural language is an essential capability for artificial intelligence. It motivates the research of image-text matching, which challenges a learning agent to establish accurate and generalizable alignment between visual and textual data, so that one can identify images or videos by text queries or vice versa.

Visual semantic embedding (VSE)~\cite{frome2013devise,kiros2014UVS,faghri2017vse++} tackles this challenge by learning a semantic embedding space, where the distance between paired visual and textual instances in the embedding space is optimized to be small.
The core idea of the VSE has three steps:
\begin{enumerate}[label={\bf {Step}~{{\arabic*}}.},leftmargin=*,topsep=0.5ex,itemsep=-1ex,partopsep=0.75ex,parsep=0.75ex,partopsep=0pt,wide, labelwidth=!,labelindent=0pt]
    \item Extract a set (or sequence) of features from data, using \textit{feature extractors} (\eg, ConvNets for visual data).
    \item Contextualize and aggregate the extracted features to project them into the joint embedding space as holistic vectors, using \textit{feature aggregators}.
    \item Compute the matching score between embeddings with a similarity metric (\eg, cosine distance).
\end{enumerate}
With the feature extractor determined, one might expect that a complex aggregator is required to achieve good results. However, we show (in \S~\ref{sec:method}) that a surprisingly simple and efficient aggregator, a carefully selected pooling function (\eg, max pooling), can surpass prior state-of-the-art VSE methods with complex aggregators~\cite{huang2018SCO,Li2019VSRN,Wehrmann2019LanguageAgnosticVE,Wu2019UniVSE,Wang2020CVSE}.
 
Such pooling functions are both simple and effective. However, searching for the optimal pooling requires extensive manual tuning and repetitive experiments (\eg, grid search) for each data modality and features, which is tedious and costly as it enumerates over a combinatorial number of configurations. This search procedure could be even more complicated when the sets of features have varying sizes.

\begin{figure*}[th!]
    \centering
    \includegraphics[width=\linewidth]{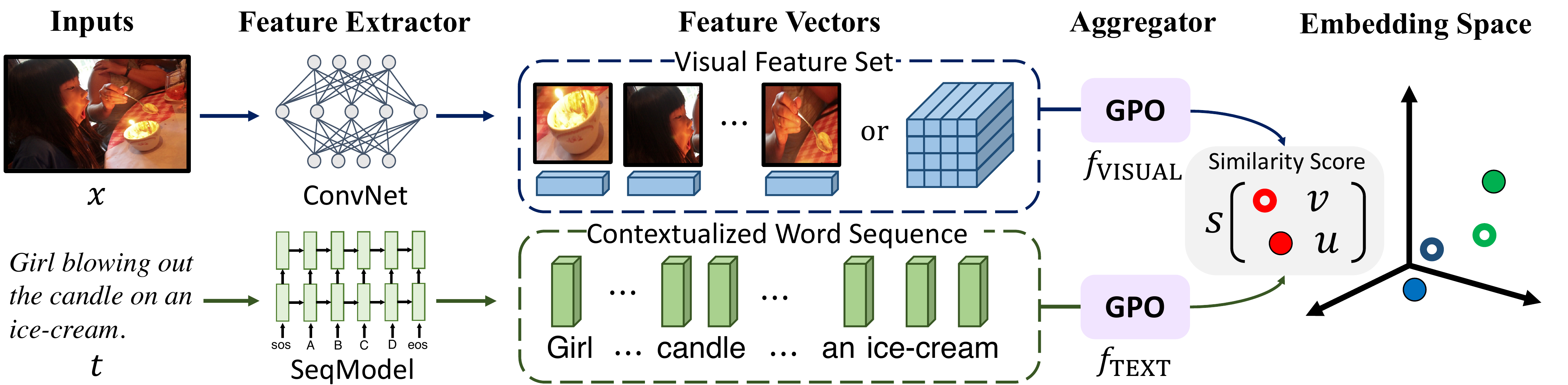}
    \vspace{-1em}
    \caption{Illustration of the standard Visual Semantic Embedding framework with the proposed pooling-based aggregator, \ie, Generalized Pooling Operator (GPO). It is simple and effective, which automatically adapts to the appropriate pooling strategy given different data modality and feature extractor, and improves VSE models at negligible extra computation cost.}
    \vspace{-1em}
    \label{fig:vse_framework}
\end{figure*}

\textit{Can we discover the best pooling strategy automatically?} 
In this paper, we propose a novel parameterized pooling operator, \emph{Generalized Pooling Operator} (GPO), to fully exploit the strengths of pooling-based feature aggregation. GPO generalizes over various pooling functions and learns to adjust itself to the best one for different data modalities and feature extractors.
Specifically, GPO learns a generator that predicts the pooling coefficients to weight the elements of sorted feature vectors, and use their weighted sum as the pooling output. The coefficient generator is instantiated as a tiny sequence model to handle variable-sized features. 
GPO learns to adapt to the optimal pooling strategy, and improve VSE models at a negligible extra computational cost.

With the proposed GPO, we build our multi-modal matching system as {\ourmethod}, which extends a standard VSE framework\cite{kiros2014UVS} by using GPO as the feature aggregators for both visual and text features. 
We train our system optimizing a margin-based triplet ranking objective similar to~\cite{faghri2017vse++}, with the online hard-negative mining.

Without bells and whistles, {\ourmethod} surpasses all previous state-of-the-art VSE-based methods on the image-text retrieval tasks, over COCO~\cite{karpathy2015deep} and Flickr30K~\cite{young2014image}. With a straightforward extension, variants of {\ourmethod} also achieve the best video-text retrieval results on two benchmark datasets, \ie,  MSR-VTT~\cite{Xu2016MSRVTTAL} and VaTeX~\cite{Wang2019VaTeXAL}. In additional experiments, we show that GPO consistently outperforms other alternative learnable poolings from the literature. To better understanding GPO, we further visualize the pooling strategy found by \ourmethod, and compare it with the one from a thorough grid search process. 

\mypara{Our contributions} are summarized as the following:
\begin{itemize}[leftmargin=*,topsep=0pt,itemsep=0pt,noitemsep]
\item We empirically find that carefully selecting simple pooling functions can outperform complex visual aggregators in prior VSE methods for image-text matching. 
\item We propose a novel Generalized Pooling Operator (GPO) that generalizes various pooling functions. It learns to automatically discover the best pooling function for image, text, and video data with various feature extractors. 
\item We build up {\ourmethod} with GPO, which achieves the new state-of-the-art performances among VSE methods on image-text and video-text retrieval.
\item We visualize the pooling strategies learned by GPO, and verify that GPO learns the best pooling strategies given the data by comparing \ourmethod with a thorough grid search over pooling functions of all modalities.
\end{itemize}
\section{Visual Semantic Embedding for \\ Multi-modal Matching}
\label{sec:framework}

We begin by revisiting the formal formulation of Visual Semantic Embedding (VSE). A VSE model (illustrated in Figure~\ref{fig:vse_framework}) leverages a visual embedding function $\mathbf{\Phi}(\vx)$ such as convolutional neural networks (\eg CNNs~\cite{He2015ResNet,Xie2016ResNeXt}), and a text embedding function $\mathbf{\Psi}(\vt)$ such as sequence models (\eg LSTMs~\cite{Hochreiter1997LSTM}, Transformers~\cite{Vaswani2017Transformer}), to compute the set of visual features and text features, respectively:
\begin{align}
    \texttt{ConvNet}(\vx) & : \vx \rightarrow \{ \vphi_{n} \}^{\cN}_{n=1}, \nonumber \\ 
    \texttt{SeqModel}(\vt) & : \vt \rightarrow \{ \vpsi_{m} \}^{\cM}_{m=1} \nonumber
\end{align}
Here the set of visual features $\{ \vphi_{n} \}^{\cN}_{n=1}$ has $\cN$ elements of convolutional local representations with $\vphi_{n} \in \mathbb{R}^{{d}_1}$. As aforementioned, the concrete form of $\vphi_{n}$ can be feature vectors of spatial grids from the feature map, object proposals~\cite{anderson2017updown}, or spatial-pyramids~\cite{he2015spatial}, depending on the feature extractor. Similarly, text features $\{ \vpsi_{t} \}^{\cM}_{m=1}$ denotes a sequence of $\cM$ contextualized word token features out of a sequence model where $\cM$ is the number of words and $\vpsi_{m} \in \mathbb{R}^{{d}_2}$. Here $d_1$ and $d_2$ are the feature dimensions. 

The output visual features $\{ \vphi_{n} \}^{\cN}_{n=1}$ 
and textual features $\{ \vpsi_{t} \}^{\cM}_{m=1}$ are then aggregated by visual and textual aggregators $f_{\textsc{visual}}(\cdot)$ and $f_{\textsc{text}}(\cdot)$, to further encode the holistic visual and text embedding ${\vv}, {\vu} \in \mathbb{R}^{{d}_3}$ as follows:
\begin{align}
    \vv = {f}_{\textsc{visual}}\Big( \{ \vphi_{n} \}^{\cN}_{n=1} \Big), \quad
    \vu = {f}_{\textsc{text}} \Big( \{ \vpsi_{m} \}^{\cM}_{m=1} \Big).  \nonumber
\end{align}
The compatibility score is then defined as the cosine similarity between $\vv$ and $\vu$, formally as:
\begin{align}
    \vcts_{(\vx, \vt)} = \frac{ \vv^\top \vu }{ \norm{\vv} \cdot \norm{\vu}} \nonumber
\end{align}
During the inference, the $\vcts_{(\vx, \vt)}$ scores are used to rank a query text against all candidate images, and the top candidate are returned as the prediction. 
We note that the inference procedure is efficient as the visual and text embedding $\vv$ and $\vu$ can be pre-computed. The pair-wise scores are then computed by matrix multiplication. 

\mypara{Learning Multi-modal Matching} 
To learn a VSE model, existing methods mostly optimize the hinge-based triplet ranking loss with online hard negative mining proposed by VSE++~\cite{faghri2017vse++}. The concrete matching objective is defined by: 
\begin{align}
    \ell_{\textsc{match}} = \sum_{(\bsx, \bst) \sim \calD} & [\alpha - \vcts_{(\bsx, \bst)} + \vcts_{(\bsx, \hat{\bst})}]^{+} \nonumber \\
    & + [\alpha - \vcts_{(\bsx, \bst)} + \vcts_{(\hat{\bsx}, \bst)}]^{+} \label{obj:match}
\end{align}
where $\alpha$ is a hyper-parameter. $(\vx, \vt)$ is a positive image-text pair in the dataset $\calD$ and $[x]^{+} \equiv \texttt{max}(0,x)$. We represent $\hat{\vt} = \texttt{argmax}_{\vt' \ne \vt}\vcts_{(\vx, \vt')}$ and $\hat{\bsx} = \texttt{argmax}_{\vx' \ne \vx}\vcts_{(\vx', \bst)}$ as the hardest negative text and image examples measured by the learned VSE model within a mini-batch. 
\section{{\ourmethod} with Generalized Pooling Operator}
\label{sec:method}

In this section, we first present an empirical finding that highlights the effectiveness of well-selected pooling function in VSE model, which motivates our methodological pursuit (\S~\ref{subsec:pooling_the_best}). We then propose our method, Generalized Pooling Operator (GPO), with a introduction of its formal definition (\S~\ref{subsec:gpo_formulation}), followed by the details of GPO's concrete model architecture (\S~\ref{subsec:gpo_implement}). Finally, we summarize our multi-modal system (\ourmethod) that leverages GPO (\S~\ref{subsec:vse_infty}).

\begin{table}[t]
    \centering
    \small
    \caption{
        {\bf Image-text retrieval results in R@1} of VSE models with different visual \textit{aggregator}, evaluated with MS-COCO 1K. See \S~\ref{subsec:retrieval_setting} for details.
    }
    \tabcolsep 5pt
    \begin{tabular}{@{}l@{\;}ccccc}
    \toprule
    & & \multicolumn{2}{c}{Region~\cite{anderson2017updown}} & \multicolumn{2}{c}{Grid~\cite{Jiang2020DefenseGridVQA}} \\
    \cmidrule(lr){3-4}\cmidrule(lr){5-6}
    {Aggregator} & \#Param & T $\rightarrow$ I & I $\rightarrow$ T & T $\rightarrow$ I & I $\rightarrow$ T \\
    \midrule
    AvgPool~\cite{faghri2017vse++} & 0 & 54.0 & 68.5 & 58.9 & 72.4 \\
    Seq2Seq~\cite{hu2019binary} & 6.3M & 58.5 & 69.9 & 61.5 & 73.3 \\
    SelfAttn~\cite{Wehrmann2019LanguageAgnosticVE,Wang2020CVSE} & 3.2M & 56.2 & 70.2 & 60.3 & 73.0\\
    GCN+AvgPool~\cite{Li2019VSRN} & 4.2M & 54.9 & 69.0 & 59.5 & 71.8 \\
    GCN+Seg2Seq~\cite{Li2019VSRN}  & 23.1M & \bf 60.7 & 72.5 & 59.5 & 71.1 \\
    \midrule
    Best Pooling Function & 0 & \bf 60.7 & \bf 74.5 & \bf 61.6 & \bf 76.3 \\
    \bottomrule
    \end{tabular}
    \label{tab:intro}
    \vspace{-0.3cm}
\end{table}

\subsection{Simple Pooling Works the Best}
\label{subsec:pooling_the_best}

As aforementioned in \S~\ref{sec:intro}, complex aggregators $f$ have been investigated in the VSE literature~\cite{huang2018SCO,Li2019VSRN,Wehrmann2019LanguageAgnosticVE,Wu2019UniVSE,Wang2020CVSE}, such as sequence-to-sequence encoder (Seq2Seq), graph convolution network (GCN), self-attention encoder (SelfAttn), \etc. However, we surprisingly find that these aggregation models with millions of parameters underperform carefully selected pooling functions. 

Table~\ref{tab:intro} highlights a comparison between different aggregators, across two widely used image feature extractors in the literature~\cite{Jiang2020DefenseGridVQA} -- \emph{Grid feature} is the feature maps from ConvNets and \emph{Region feature} is the ROI features from object detectors~\cite{anderson2017updown} (details in \S~\ref{sec:exp}). The results are reported in recall@1 for text-based image retrieval (T$\rightarrow$I) and vice versa. 
Given the candidates of Average Pooling (AvgPool), Max Pooling (MaxPool) and K-Max Pooling (K-MaxPool~\cite{kalchbrenner2014SequencewithKMax}, details in \S~\ref{subsec:gpo_formulation}) with different K, it shows that the best among them consistently outperform complex aggregators. Here, the best results for Region and Grid feature are achieved by MaxPool and K-MaxPool ($K$=20), respectively. 

\mypara{Analyses of the Empirical Findings.} 
Most complex aggregators are designed to contextualize the input features spatially, leveraging the relationship between spatial grids or regions. However, these aggregators introduce a large set of parameters in addition to the vanilla  VSE model, which causes a higher risk of over-fitting comparing to simple pooling functions. In this paper, instead of investigating why complex aggregators are suboptimal, we focus on maximizing the advantages of pooling-based aggregation.

While the optimal pooling strategy enjoys simplicity and effectiveness, searching it requires repetitive experiments over numerous configurations (\eg, different K for K-MaxPool), which is both tedious and costly.
This process can be more complicated when the feature extractor changes, or when the features have variable lengths (\eg, text).

Motivated by these, we aim for a general and plug-and-play pooling operator that generalizes over different pooling patterns (\eg, Avg, Max and K-MaxPool with arbitrary K) for variable-sized inputs, and learns to automatically adapt itself to the best strategy according to the data (\eg, image, text, video \etc) and feature extractors. We denote our proposed module as the Generalized Pooling Operator (GPO).

\subsection{Generalizing over Different Pooling Strategies}
\label{subsec:gpo_formulation}

Suppose that we have a set of $\cN$ feature vectors $\{ \vphi^{i}_{n} \}^{\cN}_{n=1}$ and our goal is to obtain a holistic vectorized embedding $\vv^i$ out from the $\cN$ elements, for each dimension $i = 1,\ldots,d_1$. Here we use the superscript $i$ to index the $i$-th dimension of the feature vector. We further denote $\texttt{max}_{k}(\cdot)$ as the operator that \emph{takes the $k$-th maximum value from an ordered list}. Then, we can formally define commonly used pooling strategies as the following:
\begin{itemize}[leftmargin=*,topsep=0pt,itemsep=1pt]
    \item \textbf{AvgPool} The average pooling computes the mean value among the $\cN$ elements, as $\vv^i = \frac{1}{\cN}\sum^{\cN}_{n=1} \vphi^{i}_{n}, \forall i$.
    \item \textbf{MaxPool} The max pooling computes the maximum value among the $\cN$ elements, as $\vv^i = \texttt{max}_1(\{ \vphi^{i}_n \}^{\cN}_{n=1}), \forall i$.
    \item \textbf{K-MaxPool} The K-max pooling computes the mean value of the top-K maximum values among the $\cN$ elements, as $\vv^i = \frac{1}{\cK}\sum^{\cK}_{k=1} \texttt{max}_{k}(\{ \vphi^{i}_n \}^{\cN}_{n=1}), \forall i$.
\end{itemize}

\mypara{Main Idea} As described above, GPO aims to generalize over various pooling strategies, so that the pooling operator can automatically find the most appropriate strategy for different features. Therefore, GPO learns to generate the pooling coefficients $\vtheta$, and the pooling is defined as a weighted sum over sorted features~\cite{Delige2019OrdinalP}: 
\begin{align}
    \vv^i = \sum^{\cN}_{k=1} \vtheta_{k} \cdot \texttt{max}_{k}\Big(\{ \vphi^{i}_n \}^{\cN}_{n=1} \Big), \forall i,\\ \text{where}\; \sum_{k=1}^{\cN} \vtheta_{k} = 1. \nonumber
\end{align}
Here, the coefficients $\vtheta$ are of the size $\cN$, with a scalar weight $\vtheta_k$ for the $k$-th maximum value among the $\cN$ elements. The constraint $\sum_{k=1}^{\cN} \vtheta_{k} = 1$ is enforced via Softmax. The parameterized pooling operator can approximate AvgPool, MaxPool, K-MaxPool with arbitrary $\cK$, and more complex pooling functions. For instance, the learned pooling strategy could weight the top-K elements unevenly, or only set non-zero values for $\vtheta_1$ and $\vtheta_N$. We visualize some learned pooling coefficients in \S~\ref{subsec:exp:visualization}.

\mypara{Learning to Generate the Pooling Coefficients} The most straightforward way to parameterize $\vtheta$ is to define it as a trainable vector, but this can only deal with the scenario where $\cN$ is a constant integer (like~\cite{Delige2019OrdinalP} did for fixed-size kernels). When the features are of variable sizes, which is common in video and text sequences, learning a fixed set of coefficients $\vtheta$ is no longer feasible. To address this issue, we propose to learn a parameterized function $g(\cdot, \cdot)$ as the \emph{coefficient generator}:
\begin{align}
    \vtheta_k = g(k, \cN), \text{where}\; k = 1,\ldots,\cN.
\end{align}
As a consequence, for each position $k$, the coefficient generator $g(\cdot, \cdot)$ outputs a coefficient $\vtheta_k$ to aggregate $\{ \vphi^{i}_n \}^{\cN}_{n=1}$.

\subsection{Implementing Generalized Pooling Operator}
\label{subsec:gpo_implement}

\begin{figure}[t]
    \centering
    \includegraphics[width=\linewidth]{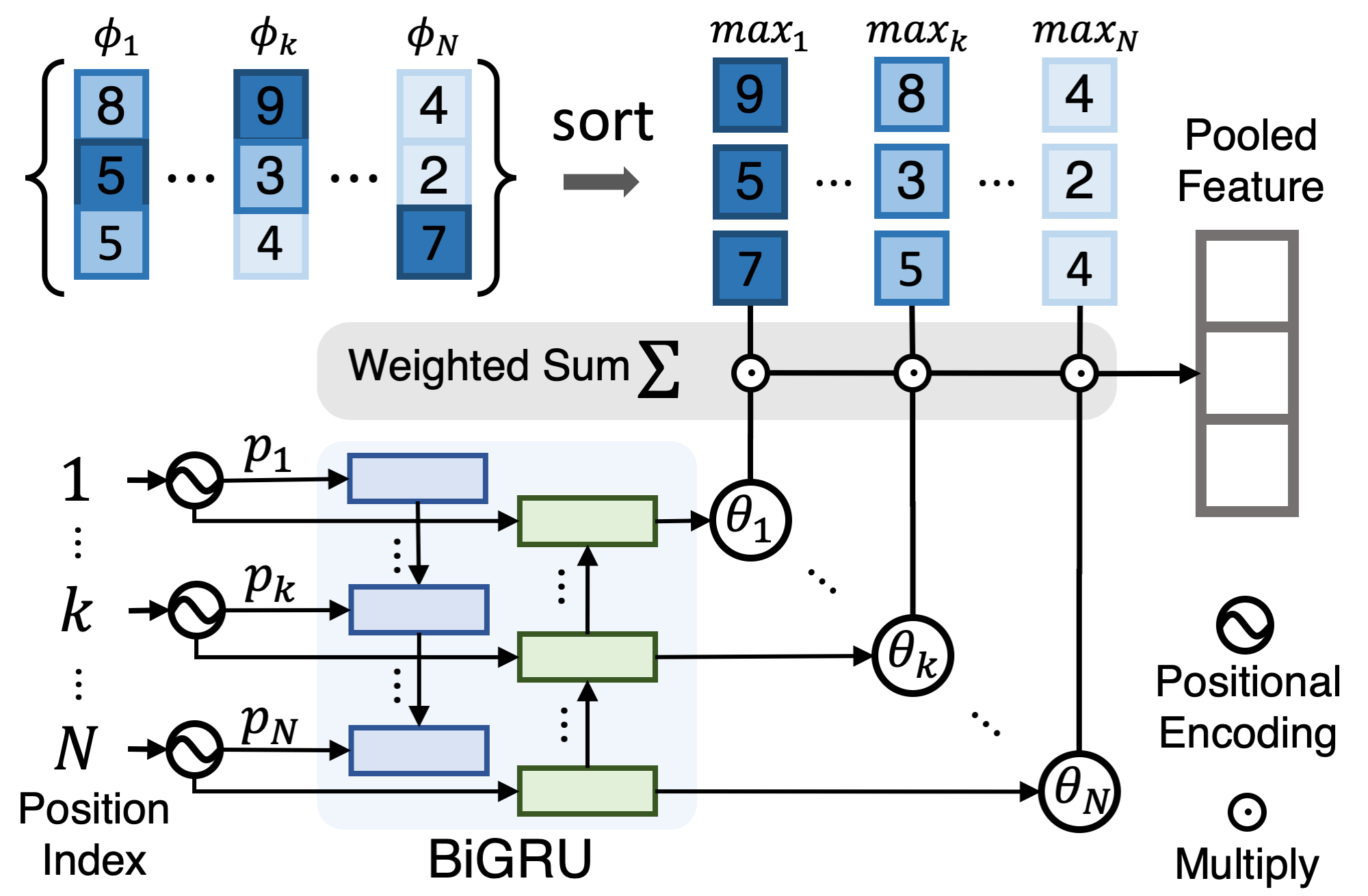}
    \caption{Detailed illustration of the GPO architecture.}
\label{fig:gpo}
\end{figure}

Now we discuss the concrete implementation of the GPO function $g(\cdot, \cdot)$. Figure~\ref{fig:gpo} provides an illustration of the architecture. There are two major components in the GPO design: (1) A positional encoding function based on trigonometric function; (2) A sequence model that takes the positional encoding sequence to generate pooling coefficients, based on bidirectional Gated Recurrent Unit (BiGRU). 

\mypara{Encoding Position}
Every position index $k$ is uniquely represented by a dense vector, such that the vector can be further transformed to $\vtheta_k$ by parameterized functions. A common approach here is to learn an embedding matrix in which row $k$ is the embedding for $k$. However, this presumes the input positions $\{1,\ldots,k,\ldots,\cN\}$ orthogonal to each other. To make more efficient use of the prior information between position indices, we adopt the positional encoding strategy used in Transformers~\cite{Vaswani2017Transformer} to vectorize positional indices:
\begin{align}
    \vp_k^i = 
    \begin{cases}
    \sin(w_j,k),& \text{when\;} i=2j \\
    \cos(w_j,k),& \text{when\;} i=2j+1
    \end{cases}
    , \forall i.
\end{align}
where $w_j=\frac{1}{10000^{2j/d_3}}$ and $d_3$ is the number of dimensions for the positional encoding. 

\mypara{Generating Pooling Coefficients with a Sequence Model}
Using the positional encoding above, we transform every position index $k$ into a dense vector $\vp_k \in \mathbb{R}^{d_3}$. Next, we learn a sequence model to produce the pooling coefficients. 
Since the size of feature set $\cN$ varies, it is necessary for the coefficient generator to be aware of the size of feature set. Therefore, we make use of a sequence-to-sequence decoder function, which takes the sequence of positional encodings $\vp=\{\vp_k\}^{\cN}_{k=1}$ as input and outputs the sequence of pooling coefficients $\vtheta=\{\vtheta_k\}^{\cN}_{k=1}$. The decoder function consists of a small BiGRU and a multi-layer perceptron (MLP):
\begin{align}
    \{\vh_k\}^{\cN}_{k=1} = \texttt{BiGRU}(\{\vp_k\}^{\cN}_{k=1}), \quad \vtheta_k = \texttt{MLP}(\vh_k)
\end{align}
Here $\vh_k$ is the output of the BiGRU at the position $k$.

\mypara{Learning Generator with Diverse Set Sizes}
To make GPO's coefficient generator $g(\cdot, \cdot)$ better approximate different pooling patterns for variable-sized inputs, we perform a data augmentation strategy to allow it observing a larger variety of feature set sizes. During the training, we randomly drop 20$\%$ inputs vectors to perturb the size of the input feature set, which we call Size Augmentation. We show in \SM that applying this strategy to both image and text effectively improve the performance of VSE models.

\subsection{Building up {\bf \ourmethod} using GPO}
\label{subsec:vse_infty}
We build up our multi-modal matching model (dubbed {\ourmethod}) by pluging GPO into the standard {VSE} framework (\S~\ref{sec:framework}). Specifically, we replace the visual and text aggregators in the standard VSE framework (\ie, AvgPool) with two GPOs. The two GPOs project the image feature vectors and text feature vectors independently into two holistic embeddings, to further compute the matching score. 
{\ourmethod} is closely related to previous VSE models. We adopt the learning framework of VSE++~\cite{faghri2017vse++} (Eq.~\ref{obj:match}), which improves early VSE models~\cite{frome2013devise,kiros2014UVS} with an additional online hard negative mining procedure. 
We refer to \S~\ref{sec:exp} for more details. 
\begin{table*}[t]
    \centering
    \small
    \caption{Image-Text Retrieval Results of VSE-based methods on COCO and Flickr30K datasets, using different visual and textual backbones (denoted by \textbf{bold section title}). $\star$: Ensemble results of two models; on IN/IN+VG/IG: Models pre-trained on ImageNet~\cite{Russakovsky2014ImageNet}, ImageNet and VisualGenome~\cite{Krishna2016VisualGenome}, or Instagram~\cite{Mahajan2018WSL}, respectively. The best and second best results (in \symtext{rsum}) are marked \textbf{bold} in \first{red} and \second{black}. We refer to the \SM for extensions of this table with more baselines and COCO 5K results.
    }
    {
        \tabcolsep 3pt
	    \begin{tabu}{@{\;} ll @{\quad} ccccccc ccccccc@{\;}}
	        \addlinespace
	        \toprule
	        {Data Split} & & \multicolumn{7}{c}{ COCO 5-fold 1$\mathrm{K}$ Test~\cite{chen2015coco-cap}} &
	        \multicolumn{7}{c}{Flickr30$\mathrm{K}$ 1$\mathrm{K}$ Test~\cite{young2014image}} \\ \cmidrule(lr){3-9} \cmidrule(lr){10-16}
	        {Eval Task} & &  \multicolumn{3}{c}{\textsc{img}~$\rightarrow$~\textsc{text}} & \multicolumn{3}{c}{\textsc{text}~$\rightarrow$~\textsc{img}} & & \multicolumn{3}{c}{\textsc{img}~$\rightarrow$~\textsc{text}} & \multicolumn{3}{c}{\textsc{text}~$\rightarrow$~\textsc{img}}  \\
	        {Method} & Feature Type & \symtext{r}{@1} & \symtext{r}{@5} & \symtext{r}{@10} & \symtext{r}{@1} & \symtext{r}{@5} & \symtext{r}{@10} & \symtext{rsum} & \symtext{r}{@1} & \symtext{r}{@5} & \symtext{r}{@10} & \symtext{r}{@1} & \symtext{r}{@5} & \symtext{r}{@10} & \symtext{rsum} \\
	        \midrule
	        \multicolumn{10}{@{\;}l}{\bf {ResNet-101 Faster-RCNN on IN+VG (\textsc{butd})~\cite{anderson2017updown} $+$ {BiGRU}}} \\[2pt]
	        LIWE~\cite{Wehrmann2019LanguageAgnosticVE}\textsubscript{2019} & Region & 73.2 & {95.5} & 98.2 & 57.9 & 88.3 & 94.5 & 507.6 & 69.6 & 90.3 & 95.6 & 51.2 & 80.4 & 87.2 & 474.3 \\
	        VSRN$\star$\cite{Li2019VSRN}\textsubscript{2019} & Region & 76.2 & 94.8 & 98.2 & {62.8} & 89.7 & 95.1 & 516.8 & 71.3 & 90.6 & 96.0 & 54.7 & 81.8 & 88.2 & 482.6 \\
	        CVSE~\cite{Wang2020CVSE}\textsubscript{2020} & Region & 69.2 & 93.3 & 97.5 & 55.7 & 86.9 & 93.8 & 496.4 & 70.5 & 88.0 & 92.7 & 54.7 & 82.2 & 88.6 & 476.7 \\
    	    Our: VSE++ & Region & 68.5 & 92.6 & 97.1 & 54.0 & 85.6 & 92.7 & 490.5 & 62.2 & 86.6 & 92.3 & 45.7 & 73.6 & 81.9 & 442.3 \\
    	    Our: \ourmethod & Region & {78.5} & {96.0} & {98.7} & {61.7} & {90.3} & {95.6} & \second{520.8} & {76.5} & {94.2} & {97.7} & {56.4} & {83.4} & {89.9} & \second{498.1} \\
    	    Our: \ourmethod & Region+Grid & {80.0} & {97.0} & {99.0} & {64.8} & {91.6} & {96.5} & \first{528.8} & {80.7} & {96.4} & {98.3} & {60.8} & {86.3} & {92.3} & \first{514.8} \\
	        \midrule
	        \multicolumn{10}{@{\;}l}{\bf {ResNet-101 Faster-RCNN on IN+VG (\textsc{butd})~\cite{anderson2017updown} $+$ {BERT}~\cite{Devlin2019BERT}}} \\[2pt]
    	    Our: VSE++ & Region & 67.9 & 91.9 & 97.0 & 54.0 & 85.6 & 92.5 & 488.9 & 63.4 & 87.2 & 92.7 & 45.6 & 76.4 & 84.4 & 449.7 \\
	        Our: \ourmethod & Region & {79.7} & {96.4} & {98.9} & {64.8} & {91.4} & {96.3} & \second{527.5} & {81.7} & {95.4} & {97.6} & {61.4} & {85.9} & {91.5} & \second{513.5} \\
	        Our: \ourmethod & Region+Grid & {82.2} & {97.5} & {99.5} & {68.1} & {92.9} & {97.2} & \first{537.4} & {85.3} & {97.2} & {98.9} & {66.7} & {89.9} & {94.0} & \first{532.0} \\
	        \midrule 
	        \multicolumn{10}{@{\;}l}{\bf {ResNeXT-101 on IG (\textsc{wsl})~\cite{Mahajan2018WSL} $+$ {BERT}~\cite{Devlin2019BERT}}} \\[2pt]
	        Our: VSE++ & Grid & 79.6 & 97.1 & 99.0 & 66.4 & 91.1 & 95.5 & 528.7 & 80.9 & 96.6 & 98.9 & 65.2 & 89.5 & 93.7 & 524.8 \\
	        Our: \ourmethod & Grid & {84.5} & {98.1} & {99.4} & {72.0} & {93.9} & {97.5} & \second{545.4} & {88.4} & {98.3} & {99.5} & {74.2} & {93.7} & {96.8} & \second{550.9} \\
	        Our: \ourmethod$\star$ & Grid & {85.6} & {98.0} & {99.4} & {73.1} & {94.3} & {97.7} & \first{548.1} & {88.7} & {98.9} & {99.8} & {76.1} & {94.5} & {97.1} & \first{555.1} \\
	        \bottomrule
	    \end{tabu}
	}
    \label{tab:vse_main}
\end{table*}

\section{Related Works}
\label{sec:related}

Existing image-text matching methods can be categorized differently based on how the \emph{cross-modal interaction} is implemented. As aforementioned, Visual Semantic Embedding (VSE)~\cite{frome2013devise,kiros2014UVS,faghri2017vse++,Li2019VSRN,Wu2019UniVSE} learns a joint embedding space, such that the compatibility score can be computed as a inner-product between the two holistic image and text vectors. Therefore, VSE relies on learning strong image and text embedding functions to obtain high-quality joint embedding space. Frome~\etal~\cite{frome2013devise} used this approach for zero-shot image recognition~\cite{lampert2013attribute,norouzi2013zero,changpinyo2016synthesized}, via matching visual embeddings with semantic word embeddings. Kiros~\etal ~\cite{kiros2014UVS} extends the idea by using bi-directional LSTMs to encode sentence as the semantic embedding. Faghri~\etal proposes VSE++, which learns with online hard-negative mining and further improves the quality of \vse models~\cite{faghri2017vse++}. VSE++ is one of the most fundamental VSE methods that use AvgPool as the feature aggregator. Beyond the above, more research along this line focused on improving the visual or text embedding function (especially the aggregator), or designing auxiliary training objectives~\cite{Socher2014GroundedCS,eisenschtat20172waynet,nam2017DAN,huang2018SCO,gu2018GXN,Song2019PolysemousVSE,Li2019VSRN,Wu2019UniVSE}.

Recently, methods using BERT models for vision-language data (V$+$L BERTs)~\cite{Lu2019ViLBERTPT,li2019visualbert,Li2019UnicoderVLAU,chen2019uniter,Li2020OscarOA,Huang2019ACMMAC} learns to perform rich cross-modal interaction, via tailored mechanisms such as (single/multi-headed) cross-attention~\cite{lee2018scan,Vaswani2017Transformer}.
These methods typically use a BERT~\cite{Devlin2019BERT} as the text feature extractor and learn additional cross-modal Transformers for rich cross-modal interactions. At the same time, these methods perform large-scale visual-linguistic pre-training with a collection of datasets with paired images and text (\eg, the Conceptual Caption dataset~\cite{sharma2018conceptual}). Comparing to this family of methods, VSE models are inferior in empirical performances as its lack of strong cross-modal interaction. However, VSE models are orders of magnitude more efficient than V$+$L BERTs in terms of cross-modal retrieval as the latter requires the huge BERT model to forward over all pairs of images and texts. In \S~\ref{subsec:exp:itm}, we show that the best \ourmethod can attain a close image-text matching performance to the best V$+$L BERT method while being much faster in large-scale multi-modal retrieval.
\section{Experiments}
\label{sec:exp}

We conduct experiments to validate \ourmethod on image-text (\S~\ref{subsec:exp:itm}) and video-text matching (\S~\ref{subsec:exp:vtm}). We compare GPO with alternative poolings in \S~\ref{subsec:exp:comparison}, and analyze the learned GPO in \S~\ref{subsec:exp:visualization}. We refer to the \SM for complete experimental details and more ablation studies.

\subsection{Multi-modal Retrieval Experiments}
\label{subsec:retrieval_setting}

Multi-modal retrieval is typically evaluated using the metric of recall at K (\symtext{R}{@K}), with $K = \{1, 5, 10\}$. We follow~\cite{Chen2020IMRAMIM,Wu2019UniVSE} to use $\symtext{rsum}$, which is defined as the sum of recall metrics at $K = \{1, 5, 10\}$ of both I$\rightarrow$T (I2T) and T$\rightarrow$I (T2I) retrievals, as a summarizing metric to gauge retrieval model's overall performances. In all experiments, we set the dimensions of the positional encoding and BiGRU to be 32. Therefore, GPO has 0.1$M$ parameter in total, which is less than $1\%$ of the entire model. 

\subsubsection{Image-text Retrieval}
\label{subsec:exp:itm}

\mypara{Setup} For image-text retrieval, we perform experiments on MS-COCO~\cite{Lin2014MicrosoftCC,chen2015coco-cap} and Flickr30K~\cite{young2014image} over various feature extractors. Each image of these two datasets is associated with five text descriptions. COCO contains 123,287 images, we use the data split of~\cite{karpathy2015deep,faghri2017vse++,lee2018scan} where there are 113,287 training images, 5000 test images, and 5000 validation images. Flickr30K contains 31,000 images, we also use the same data split as~\cite{faghri2017vse++}, where there are 29,000 training images, 1000 test images, and 1000 validation images. COCO results are reported in 5K and 1K, where the 1K results are averaged over the five 1K data folds. 
The image feature extractors are categorized into \emph{Region feature} and \emph{Grid feature} following the naming convention in~\cite{Jiang2020DefenseGridVQA}, where grid feature represents the feature maps from a CNN, and region feature represents object-level features from a detector.

\mypara{Implementation Details}
The dimension of the joint embedding space is 1024. We use pre-extracted object features~\cite{anderson2017updown} as the region feature (\textsc{butd} feature). For grid feature, the CNN backbone is fine-tuned, and we increase the resolution of input images to 512$\times$512 as suggested by~\cite{Jiang2020DefenseGridVQA}. We experiment with two different CNNs: 
(1) ResNet-101 of Faster-RCNN~\cite{ren2015faster} pre-trained on ImageNet and Visual Genome (\textsc{butd})~\cite{anderson2017updown} and (2) ResNeXT-101(32$\times$8d)~\cite{Xie2016ResNeXt} pre-trained on Instagram (\textsc{wsl})~\cite{Mahajan2018WSL}.
Meanwhile, we use either BiGRU or BERT-base as the text feature extractor. We refer to the \SM for full training details and more results. 

\begin{table}[t]
    \centering
    \small
	\tabcolsep 3pt
    \caption{
    	Comparison between variants of \ourmethod and V$+$L BERTs. All methods uses BERT-base. $\star$: ensemble results of two models. R/G in parenthesis represents Region/Grid features. 
    }
    \begin{tabular}{@{\;}l@{}c@{\;}c@{\;\;}cccc}
        \addlinespace
        \toprule
        {Data Split} & & & \multicolumn{4}{c}{COCO 5$\mathrm{K}$ Test~\cite{chen2015coco-cap}} \\
        {Eval Task}  & & & \multicolumn{2}{c}{ \textsc{img}~$\rightarrow$~\textsc{text}} & \multicolumn{2}{c}{ \textsc{text}~$\rightarrow$~\textsc{img}} \\
        {Method} & {Pretrain} & {CNN} & \symtext{r}{@1} & \symtext{r}{@5} & \symtext{r}{@1} & \symtext{r}{@5} \\
        \midrule
        Vi\textsc{lbert}\cite{zhang2020learning} & \cmark & \textsc{butd} & 53.5 & 79.7 & 38.6 & 68.2 \\
        Vi\textsc{lbert} DG\cite{zhang2020learning} & \cmark & \textsc{butd} & 57.5 & 84.0 & 41.8 & 71.5 \\
        \textsc{Unicoder} VL\cite{Li2019UnicoderVLAU} & \cmark & \textsc{butd} & 62.3 & 87.1 & 46.7 & 76.0 \\
        \textsc{Uniter}\cite{chen2019uniter} & \cmark & \textsc{butd} & 64.4 & 87.4 & 50.3 & 78.5  \\
        \textsc{Oscar}\cite{Li2020OscarOA} & \cmark & \textsc{butd} & 70.0 & 91.1 & 54.0 & 80.8  \\
        \midrule
        \multicolumn{6}{@{}l}{\bf {Our Methods} } \\
         \ourmethod(R)   & \xmark & \textsc{butd} & 58.3 & 85.3 & 42.4 & 72.7 \\
         \ourmethod(R+G) & \xmark & \textsc{butd} & 62.5 & 87.8 & 46.0 & 75.8 \\
         \ourmethod(G)   & \xmark & \textsc{wsl} & 66.4 & 89.3 & 51.6 & 79.3  \\
         \ourmethod(G)~${\star}$ & \xmark & \textsc{wsl} & 68.1 & 90.2 & 52.7 & 80.2 \\
         \bottomrule
    \end{tabular}
    \label{tab:cross}
\end{table}
\begin{table*}[tbh!]
    \centering
    \small
	\tabcolsep 6pt
    \caption{
    	Results on video-text retrieval benchmarks. $\infty$: Methods modified to using the GPO to aggregate frame and word features. 
    }
    \vspace{-1em}
    {
    \subfloat[\bf \small MSR-VTT Video-Text Retrieval~\cite{Xu2016MSRVTTAL}]
    {
    \begin{tabular}{@{\;}lccccccc}
    	        \addlinespace
    	        \toprule
    	        \multirow{2}{*}{Method} & \multicolumn{3}{c}{ \textsc{video}~$\rightarrow$~\textsc{text}} & \multicolumn{3}{c}{ \textsc{text}~$\rightarrow$~\textsc{video}} &  \\
    	        & \symtext{r}{@1} & \symtext{r}{@5} & \symtext{r}{@10} & \symtext{r}{@1} & \symtext{r}{@5} & \symtext{r}{@10} & \symtext{rsum} \\
    	        \midrule
    	        VSE++\cite{faghri2017vse++} & 14.4 & 34.1 & 45.6 & 8.3 & 24.0 & 34.1 & 160.5 \\ 
    	        VSE$\infty$ & 16.0 & 38.6 & 50.2 & 8.7 & 25.3 & 35.9 & \first{174.7} \\
    	       \midrule
    	        HGR~\cite{chen2020fineHGR} & 15.0 & 36.7 & 48.8 & 9.2 & 26.2 & 36.5 & 172.4 \\ 
    	        HGR$\infty$ & 15.0 & 39.0 & 51.7 & 9.1 & 25.9 & 36.3 & \first{177.0} \\
    	    \bottomrule
        \end{tabular}
    }
    \subfloat[\bf \small VATEX Video-Text Retrieval~\cite{Wang2019VaTeXAL}]
    {
    \begin{tabular}{ccccccc@{\;}}
    	        \addlinespace
    	        \toprule
    	          \multicolumn{3}{c}{ \textsc{video}~$\rightarrow$~\textsc{text}} & \multicolumn{3}{c}{ \textsc{text}~$\rightarrow$~\textsc{video}} &  \\
    	         \symtext{r}{@1} & \symtext{r}{@5} & \symtext{r}{@10} & \symtext{r}{@1} & \symtext{r}{@5} & \symtext{r}{@10} & \symtext{rsum} \\
    	        \midrule
    	         47.8 & 78.6 & 86.2 & 34.7 & 71.3 & 81.7 & 400.3  \\
    	         51.2 & 78.7 & 86.3 & 34.2 & 71.6 & 81.9 & \first{403.9} \\ 
    	         \midrule
    	         48.9 & 79.1 & 87.9 & 35.6 & 73.5 & 83.4 & 408.4 \\
    	         51.0 & 78.8 & 87.7 & 37.3 & 73.4 & 82.4 & \first{410.6} \\
    	    \bottomrule
    	    
        \end{tabular}
    }
    }
    \label{tab:video-text}
\end{table*}

\mypara{Main Results} Table~\ref{tab:vse_main} compares \ourmethod with VSE baselines over different feature extractors. VSE++ is the fundamental VSE method as described in \S~\ref{sec:related}, we re-implement it (denoted as \emph{Our: VSE++}) and apply it on latest feature extractors (\eg, \textsc{butd} image features, BERT, \etc). The major difference to its original implementation is the input image size for grid feature. LIWE~\cite{Wehrmann2019LanguageAgnosticVE}, VSRN~\cite{Li2019VSRN}, and CVSE~\cite{Wang2020CVSE} are state-of-the-art VSE methods proposed in recent two years (we compare with more baselines in the \SM). We use numbers directly from original papers except for CVSE, for which we re-run the official code after removing unfair additional label inputs and fixing its 1K evaluation setting (details in \SM).
\emph{Region+Grid} means training two separate models with region and grid feature and averaging their similarity outputs. Over all three combinations of feature extractors, \ourmethod outperforms the baselines \emph{without using complicated aggregator}. Besides, \ourmethod with WSL+BERT as feature extractors achieves the best empirical results, improving over the second best feature extractors by a large margin. \ourmethod is better than the baselines in both \emph{performance} and \emph{simplicity}. We present the COCO 5K Test results, and results with additional feature extractors in the \SM.

\begin{figure}[t]
    \centering
    \includegraphics[width=0.9\linewidth]{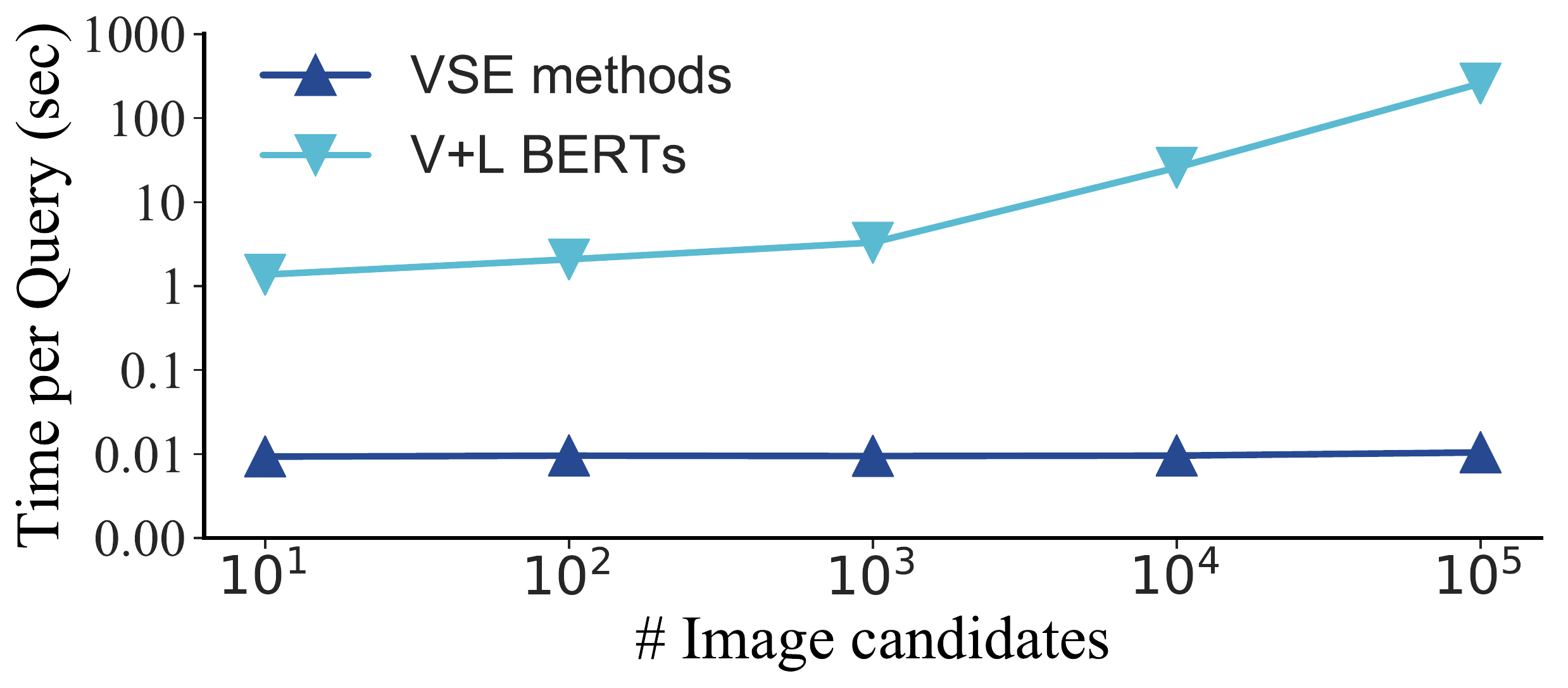}
    \caption{We compare single GPU inference time for text-based image retrieval (lower the better). VSE methods are much faster than V+L BERTs, especially when number of images grows large.}
\label{fig:retireval_speed}
\end{figure}

\mypara{Comparing \ourmethod with V$+$L BERTs}
We further compare \ourmethod with state-of-the-art V$+$L BERTs in Table~\ref{tab:cross}. We report results on COCO 5K as the 1K results reported by V$+$L BERTs is computed on the first 1K fold, instead of the average result over the five 1K folds. 
Without large-scale V+L pre-training, our \ourmethod(R+G) is no worse than three out of five V$+$L BERTs using the same feature extractors. By using the WSL CNN to compensate for the lack of pre-training, \ourmethod further outperforms UNITER and gets very close to OSCAR~\cite{Li2020OscarOA}, which is the current best V$+$L BERT. This is a promising result since VSE models by design do not have~\emph{any fine-grained cross-modal interaction} as V+L BERTs (see \S~\ref{sec:related}). Meanwhile, VSE methods are orders of magnitude faster for large-scale multi-modal retrieval as the holistic embeddings can be pre-computed or indexed~\cite{johnson2019Faiss}, and matrix multiplication is all we need to compute the compatibility score. To demonstrate this, we perform an additional text-to-image retrieval experiment with increasing size of image candidates, and visualize the model's inference time in Figure~\ref{fig:retireval_speed}. When the number of image candidates is small, we observe that VSE is a hundred time faster than V$+$L BERT. As the number of image candidates grows, the gap of time cost increases almost quadratically. \ourmethod fully exploits existing feature extractors and pushes the performance of VSE-based methods to a new height, which have significant impact in real-world problems such as image search with text query.

\begin{table}[tbh!]
    \centering
    \small
	\tabcolsep 2pt
    \caption{
    	Evaluations on COCO 5K test set with Crisscrossed Caption (CxC). All models are trained on the COCO dataset. 
    }
    {
    \begin{tabular}{@{\;}lcccccccccccccc}
    	        \addlinespace
    	        \toprule
    	        \multirow{2}{*}{Method} &
    	        \multicolumn{2}{c}{ \textsc{i}~$\rightarrow$~\textsc{t}} & \multicolumn{2}{c}{ \textsc{t}~$\rightarrow$~\textsc{i}} & \multicolumn{2}{c}{ \textsc{t}~$\rightarrow$~\textsc{t}} & \multicolumn{2}{c}{ \textsc{i}~$\rightarrow$~\textsc{i}}\\
    	        & \symtext{r}{@1} & \symtext{r}{@5} & \symtext{r}{@1} & \symtext{r}{@5} & \symtext{r}{@1} & \symtext{r}{@5} & \symtext{r}{@1} & \symtext{r}{@5} \\
    	        \midrule
    VSRN~\cite{Li2019VSRN} & 52.4 & 81.9 & 40.1 & 71.1 & 41.0 & 64.8 & 44.2 & 76.7 \\
    DE~\cite{parekh2020crisscrossed} & 55.9 & 84.2 & 41.7 & 72.3 & 42.6 & 64.9 & 38.5 & 73.6 \\
    \ourmethod (\textsc{butd}) & 60.6 & 87.4 & 46.2 & 76.3 & 45.9 & 68.7 & 44.4 & 78.3 \\
    \ourmethod (\textsc{wsl}) & 67.9 & 90.6 & 53.6 & 81.1 & 46.7 & 69.2 & 51.3 & 83.2 \\
    \bottomrule 
    \end{tabular}
    }
    \label{tab:cxc}
\end{table}

\mypara{Evaluating \ourmethod with Crisscrossed
Captions} We evaluate our best models (with BERT and Grid features on either \textsc{butd} or \textsc{wsl} backbones) on the Crisscrossed Captions(CxC)~\cite{parekh2020crisscrossed} extension of COCO, which evaluates image-text matching systems more holistically with additional intra-modal and inter-modal semantic similarity annotations. Table~\ref{tab:cxc} shows that our model can significantly outperform the baseline for both inter-modality and intra-modality (on \textsc{butd} features). Moreover, \ourmethod with \textsc{wsl} feature can further boost the performances.
\subsubsection{Video-text Retrieval}
\label{subsec:exp:vtm}

\mypara{Setup} We evaluate our method on two video datasets: MSR-VTT~\cite{Xu2016MSRVTTAL} and VATEX~\cite{Wang2019VaTeXAL}. MSR-VTT contains 10,000 videos while each video has 20 text descriptions, and we use the standard split with 6573 videos for training, 2990 for testing and 497 for validation. VATEX contains 25,991 videos for training, 6000 for testing and 3000 for validation, and the 10 English descriptions for each video are used in the experiments. We splits the original validation set into new validation and testing set, each with 1500 videos, as~\cite{chen2020fineHGR}. 

\mypara{Implementation Details} We use ResNet-152 pre-trained on ImageNet to extract frame features for MSR-VTT and use the official I3D feature for VATEX. All implementations are based on the official code of the video-text matching method HGR~\cite{chen2020fineHGR}, and we re-train all models. BiGRU is the text backbone for all experiments and the VSE setting is similar to~\ref{subsec:exp:itm} except that visual features are frame-level video features. Complete details are in the \SM.

\mypara{Main Results} Table~\ref{tab:video-text} presents the effectiveness of {\ourmethod} on video-text matching. VSE++ for video-text matching is an extension of the image-text version. HGR~\cite{chen2020fineHGR} is the current state-of-the-art method, which employs hierarchical matching strategies. By replacing the AvgPool on frames and text with GPO, \ourmethod clearly outperforms VSE++ in terms of RSUM. Additionally, we change the pooling function in the global-matching branch of HGR~\cite{chen2020fineHGR} with GPO (denoted as HGR$\infty$), and get consistent improvements. 

\begin{figure}[th!]
    \centering
    \includegraphics[width=\linewidth]{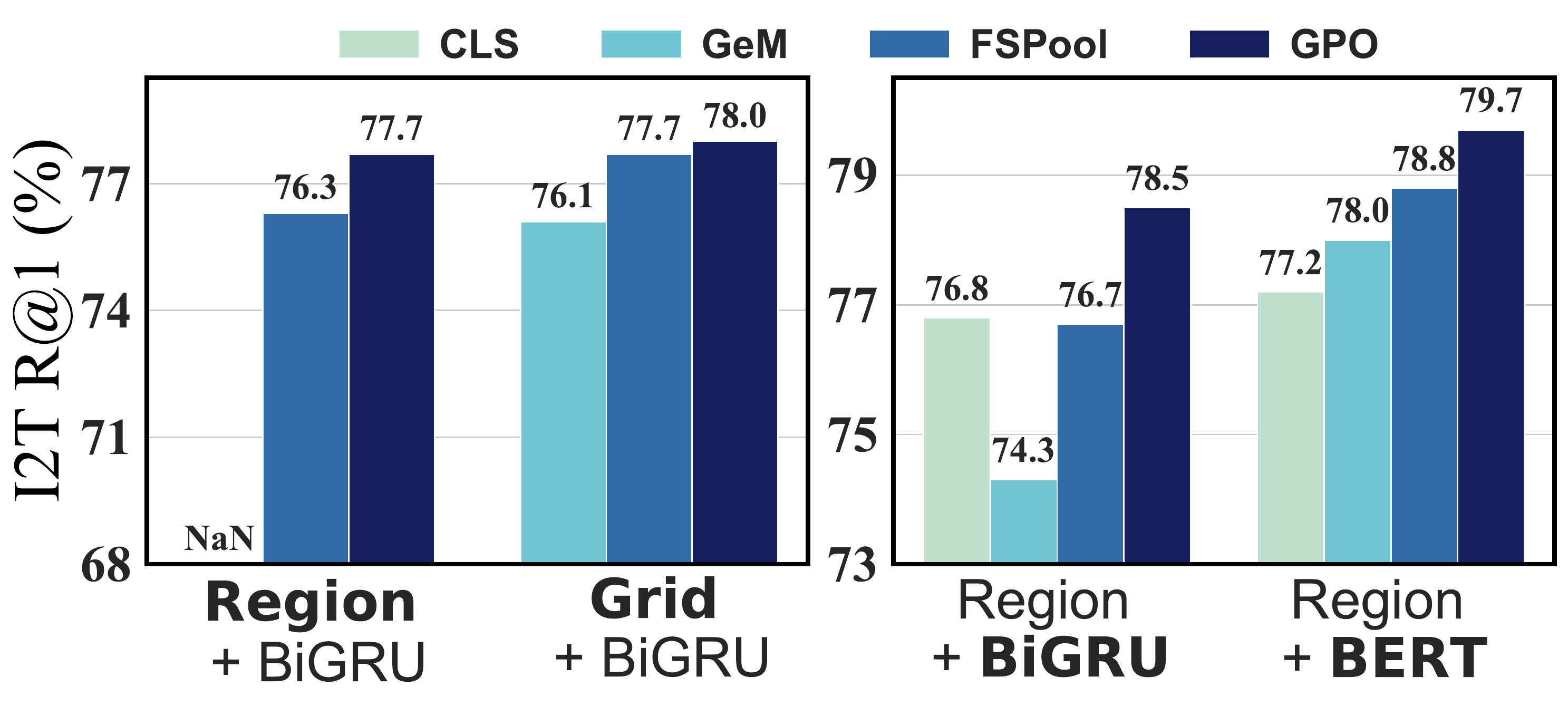}
    \vspace{-1em}
    \caption{
    \textbf{Left figure} studies different aggregators on two visual features, with AvgPool as the text aggregator for BiGRU. \textbf{Right figure} studies different aggregators on two textual features, while using {GPO} as the visual aggregator for the region features.
    }
    \vspace{-1em}
\label{fig:compare_pool}
\end{figure}

\begin{figure*}[t]
    \centering
    \includegraphics[width=\linewidth]{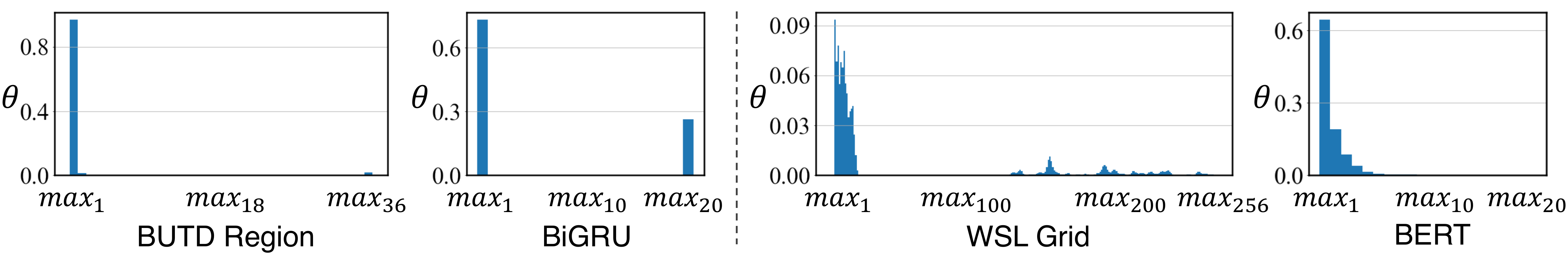}
    \vspace{-0.5em}
    \caption{
    Visualization of pooling coefficients learned by GPO. The left and right figures are the VSE models on ``{BUTD} Region$+$BiGRU'' and  ``{WSL} Grid$+$BERT'' features, respectively.
    }
    \vspace{-0.5em}
    \label{fig:viz_gpo}
\end{figure*}

\begin{figure}[t]
    \centering
    \includegraphics[width=\linewidth]{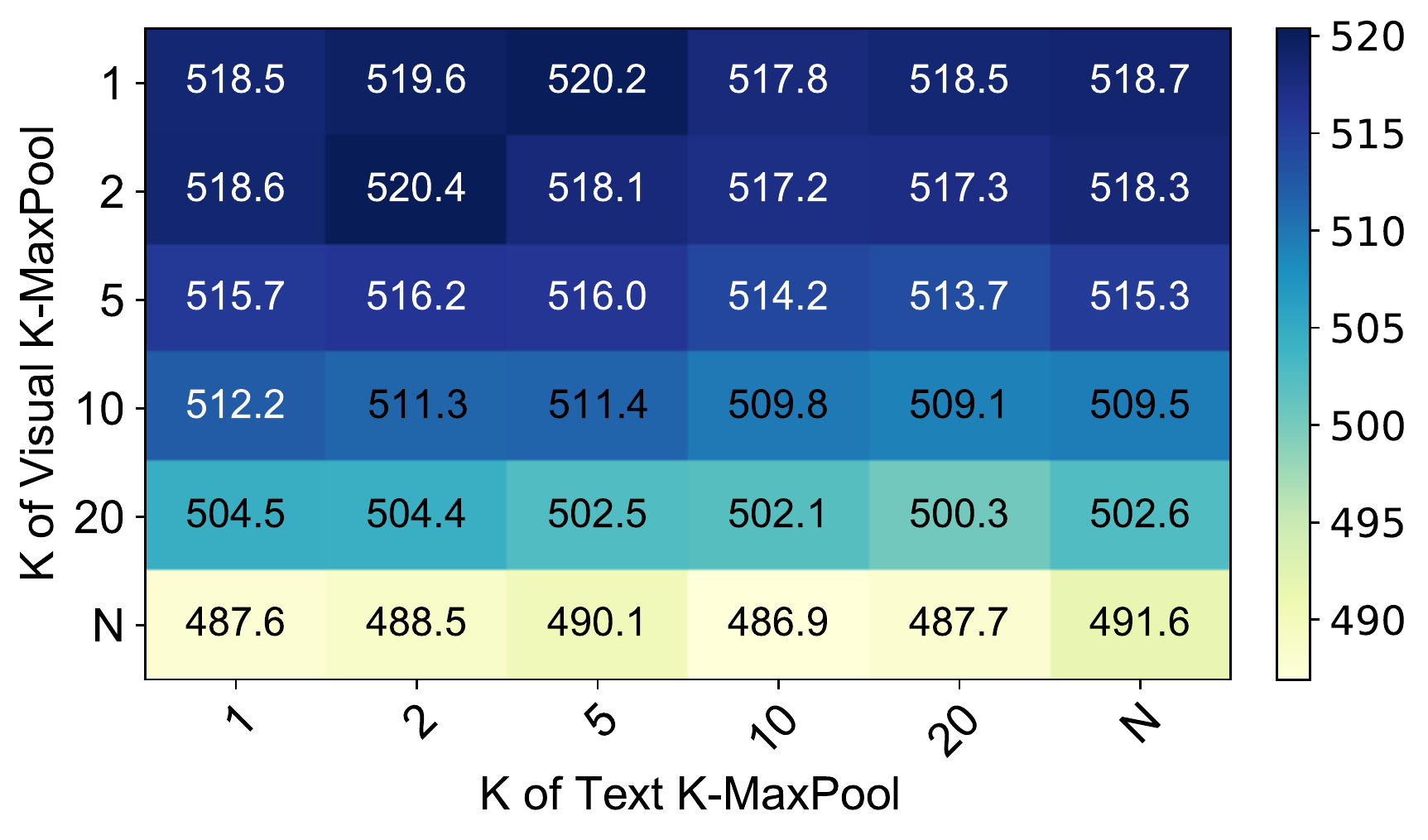}
    \vspace{-0.5em}
    \caption{Illustrating the grid search process using K-MaxPool with various K, on \textsc{butd} region and BiGRU features. The results are the $\symtext{rsum}$ values of COCO 5-fold 1K evaluation.}
    \label{fig:grid_search}
\end{figure}

\subsection{Comparing GPO to Alternative Poolings}
\label{subsec:exp:comparison}
We compare GPO with several representative learnable pooling methods across four combos of visual and text feature extractors. GPO's Size Augmentation is used in all cases for fair comparison. The baselines include:
\begin{itemize}[leftmargin=*,nolistsep]
    \item \textbf{Generalized Mean Pooling (GeM)~\cite{Radenovi2019FineTuningCIGeM}}, an adaptive pooling function with a single trainable parameter, and is popular in image search literature;
    \item \textbf{Feature-sorting Pooling (FSPool)~\cite{Zhang2020FSPoolLS}}, a learnable pooling that handles variable-sized inputs by interpolating a fixed-size learnable vector, which was proposed to encode sets in permutation-invariant manner. FSPool generates different pooling coefficients for each feature dimension.
    \item \textbf{CLS} token based aggregation, which is widely used for aggregating text features in the BERT models~\cite{Devlin2019BERT}. For BiGRU, we simply take the feature of the first token for the CLS aggregation.
\end{itemize}
Figure~\ref{fig:compare_pool} presents the comparison in R@1 of COCO I2T, we skip T2I results since the conclusions are the same. In the left part of visual pooling, FSPool is close to GPO on grid feature, but much worse on region feature. We note that the official GeM implementation is not numerical stable and it causes gradient explosion when training with region feature and BiGRU. In the right of Figure~\ref{fig:compare_pool}, we vary over different textual pooling strategies, with BiGRU or BERT being the text feature extractor. Again, GPO outperforms all alternative pooling methods. It is worth noting that BERT's default CLS aggregator is far from being optimal in the context of multi-modal matching. Above all, GPO is the best pooling strategy on various combinations of features, and can serve as a plug-and-play per-modality feature aggregator.

\subsection{Visualizing and Understanding GPO}
\label{subsec:exp:visualization}

To better understand the pooling patterns learned by GPO, we visualize the learned pooling coefficients of GPO in Figure~\ref{fig:viz_gpo}. On \textsc{butd} region feature, GPO approximates MaxPool, which is consistent with the observation in \S~\ref{sec:method}. On grid feature, the coefficients are less regular, but large position indices take up most large values. We additionally observe that GPO generates non-zero coefficients for the maximum and minimum values of BiGRU features, which goes beyond the pattern of K-MaxPool. The learned pooling strategy for BERT is close to MaxPool. 

\mypara{Comparing GPO against Grid Search.} We recall that the main motivation of GPO is to fully exploit the advantages of simple pooling functions but eliminate the repetitive manual experiments for seeking the best pooling hyperparameter. To verify how GPO address this challenge, we conduct a manual grid search over K-MaxPool with different K values for image-text matching with \textsc{butd} region and BiGRU features. GPO's Size Augmentation is used here for fair comparison as it improves performance (see \SM for details). As shown by Figure~\ref{fig:grid_search}, the best $\symtext{rsum}$ given by the grid search is 520.4, which is slightly worse than the 520.8 of the corresponding GPO entry in Table~\ref{tab:vse_main}, which means that GPO successfully refrains us from the costly repetitive search. Note that GPO generates a pooling strategy beyond K-MaxPool for BiGRU (Figure~\ref{fig:viz_gpo}), although it does not make it significantly better than the best-selected K-MaxPool.

Figure~\ref{fig:grid_search} shows that the best combination of pooling functions for visual and text modalities are entangled with each other. For instance, the best textual pooling function varies when the visual pooling function is changed. Therefore, a $n\times n$ search is necessary to find the optimal combinations of K, where $n$ is the number of grids for each modality. This could become worse when the visual feature includes multiple feature extractors (\eg, region$+$grid), as the search complexity can further become $O(n^{3})$.

In summary, GPO keeps the effectiveness and efficiency of best-selected pooling functions, and avoids the annoying grid search process. GPO can serve as a plug-and-play aggregation module to improve VSE models. 

\section{Conclusion}
\label{sec:conclusion}
In this paper, we propose the Generalized Pooling Operator (GPO), which learns to automatically adapt itself to the best pooling strategy for different data and feature backbone. As a result, we build up our \ourmethod by extending the standard VSE model with GPO as the feature aggregators. {\ourmethod} outperforms previous VSE methods significantly on image-text retrieval benchmarks across popular feature extractors. We further demonstrate that our VSE model achieves comparable image-text matching performances to vision$+$language BERT models, without visual-linguistic pre-training. Comprehensive ablation experiments confirm that GPO discovers proper pooling strategies. With simple adaptations, variants of {\ourmethod} further demonstrate effectiveness by achieving the new state of the art on two video-text retrieval datasets.

\clearpage

\appendix

\begin{center}
    \large \textbf{Appendix}
\end{center}

\noindent In this \SM, we provide implementation details and experiments omitted in the main text. The content is organized as follows:
\begin{enumerate}[leftmargin=*,itemsep=1pt]
    \item Additional implementation details, including model settings and training setups for different experiments.
    \item More experiments and results, including
    \begin{itemize}[leftmargin=*,topsep=0pt,itemsep=0pt,noitemsep]
        \item Ablation studies for better understanding GPO's effects in \ourmethod.
        \item Extensions of Table~2 of the main text, with full COCO 5K evaluation results, more combinations of feature backbones, and more related baselines. 
        \item A synthetic experiment for further verifying the design choices of GPO, as well as motivating the Size Augmentation.
        \item Experiments for exploring more complex variants of GPO (\ie, data-dependent pooling and per-dimensional pooling), which provide potential explanations for the failure of complex feature aggregators.
    \end{itemize}
\end{enumerate}

\section{Additional Implementation Details}
\subsection{Image-text Matching}

\mypara{Model settings}
The dimensionality $d_3$ of the joint embedding space is 1024 for all experiments. Note when ${d_1}\neq{d_3}$ or ${d_2}\neq{d_3}$, MLPs are applied before GPO to transform the dimensionality of features. The CNN backbones used in our experiments are either ResNet~\cite{He2015ResNet} or ResNeXt101~\cite{Xie2016ResNeXt}, thus $d_1$ is 2048. For text backbones, we set $d_2=1024$ for BiGRU, and pre-trained BERT has default $d_2$ being 768. We convert the officially released \textsc{butd} CNN from Caffe to Pytorch for running experiments related to \emph{Grid} feature. When using ~\emph{Grid} feature, the dimensionality transformation only uses a single linear layer since the image branch already has a massive CNN. When using ~\emph{Region} feature, the transformation uses a two-layer MLP with residual link.

\mypara{Training details}
The VSE models for image-text matching are trained with AdamW optimizer with weight decay factor 10e-4. The batch size is always 128 and the margin $\alpha$ of the triplet ranking loss is 0.2. The initial learning rate is 5e-4 while different model components have different learning rate multiplier: (1) CNNs pre-trained on ImageNet: 0.1; (2) \textsc{butd} or \textsc{WSL} CNNs: 0.01; (3) BERT: 0.1. The models are trained for 25 epochs and the learning rate decays by a factor of 10 for the last 10 epochs.  When fine-tuning CNN backbones (\ie, for \emph{Grid} feature), we fix the running statistics of the Batch Normalization layers during the training.

Two warm-up strategies are used during training: (1) At the first epoch, the parameter of the ConvNet is not trained (only for \emph{Grid} feature), and the triplet ranking loss uses all negative examples in the batch instead of using only the hardest negative example; (2) Starting from the second epoch, all parameters are trained end-to-end with Eq.~1 of the main text, and linear learning rate warmup is used.

All experiments are implemented with PyTorch v1.2.0 and run on Tesla V100 PCI-E GPU.

\mypara{Fixes of CVSE~\cite{Wang2020CVSE} for fair comparison}
We notice that the official implementation of CVSE~\cite{Wang2020CVSE} shows two unfair experimental setups compared to other image-text matching methods in the literature: 
\begin{enumerate}[leftmargin=*,itemsep=1pt]
\item The ``concept labels''(\ie, words with semantic meaning) are provided as extra input for the model, which is one of the core contributions of CVSE. However, for each text input, its ``concept labels'' actually come from \emph{all five ground-truth captions} (in COCO and Flickr30k, each image is associated with five captions). This is an unfair information leak compared to other methods, as it makes the model leverage information from five captions to match one image. The valid ``concept labels'' of each caption should only contain labels from itself.
\item Previous image-text matching methods report COCO 1K results by averaging over five 1K data folds of the test set, but CVSE is only evaluated on the first 1K fold.  
\end{enumerate}
We use the official released code\footnote{\url{https://github.com/BruceW91/CVSE}} to re-train and re-evaluate the model while fixing the above two setups, and report the results in the main paper.

\subsection{Video-text Matching}

\mypara{Model settings \& Training details}
We follow the official implementation of HGR~\cite{chen2020fineHGR}\footnote{\url{https://github.com/cshizhe/hgr_v2t}} in the video-text matching experiments and plug in our GPO as the video and text feature aggregator. The same as the image-text matching experiments, the dimension of the joint embedding space is 1024 and the margin $\alpha$ of triplet ranking loss is 0.2. All video-text models are trained for 35 epochs with batch size 128. The initial learning rate is 1e-4, and the learning rate decays by a factor of 10 for the last 10 epochs. 

We found that re-running the video-text VSE++ baseline in the official HGR code produces obviously higher results compared to those reported in the original paper, thus we used our re-running results to compare VSE++ and VSE$\infty$.

\subsection{Setups of Figure~3}
To measure the speed of text-based image retrieval with VSE and V$+$L BERT in Figure~3, we first pre-compute all features/embeddings that are not conditioned on the text query (\ie, holistic embeddings for VSE, and \textsc{butd} region features for V$+$L BERT). Then we compute the similarity scores between the text query and all image candidates. For VSE models, all similarity scores are calculated with a large matrix multiplication, while for V$+$L BERT, we need to forward the BERT model for $n$ times where $n$ is the number of image candidates. We tune the batch size so that the GPO memory is fully utilized.

\section{Additional Experiments and Results}

\subsection{More Ablation Studies}
\mypara{(1) Do we need GPO for both visual and text features?}
\begin{table}[t]
\centering
\small
\caption{Variants of \ourmethod on \textsc{butd} region feature and BERT, with different aggregator combintations.}
\vspace{0.5em}
\begin{tabular}{cc@{\quad} cc @{\quad} cc}
\toprule
 \multicolumn{2}{l}{Data Split}  & \multicolumn{4}{c}{ COCO 5-fold 1$\mathrm{K}$ Test~\cite{chen2015coco-cap}} \\
 \multicolumn{2}{l}{Eval Task} & \multicolumn{2}{c}{\textsc{img}~$\rightarrow$~\textsc{text}} & \multicolumn{2}{c}{\textsc{text}~$\rightarrow$~\textsc{img}} \\
 {Visual Aggr.} & Text Aggr. & \symtext{R}{@1} & \symtext{R}{@5} & \symtext{R}{@1} & \symtext{R}{@5} \\
\midrule
AvgPool & CLS & 67.7 & 92.6 & 54.8 & 85.1  \\
AvgPool & GPO & 71.7 & 93.8 & 57.5 & 87.6  \\
GPO & CLS & 73.8 & 94.8 & 59.2 & 88.0 \\
GPO &  GPO & 79.7 & 96.4 & 64.8 & 91.4 \\
\bottomrule
\end{tabular}
\label{tab:gpo_ablation}
\end{table}
In Table~\ref{tab:gpo_ablation}, we use GPO to replace the standard pooling function for \textsc{butd} image region feature and BERT text feature (\ie, AvgPool and CLS). The comparisons confirm that GPO is effective for either modality alone, using GPO for both modalities can further improve the results by a large margin. 

\mypara{(2) Effectiveness of Size Augmentation}
Table~\ref{tab:aug} shows the effectiveness of Size Augmentation. The random dropping of either visual and text inputs boost the multi-modal retrieval performance. A synthetic experiment in \SM provides more motivations for this augmentation strategy. Surprisingly, we find this strategy also improves the baseline VSE++~\cite{faghri2017vse++}, potentially as a regularization method. State-of-the-art video-text matching model~\cite{chen2020fineHGR} uses a similar strategy, feature dropout, which adds a dropout layer for input features. The difference is that it randomly set feature values to 0, while Size Augmentation randomly drops entire elements from the feature set.
\begin{table}[t]
\centering
\tabcolsep 4pt
\small
\caption{The Size Augmentation's effect on different modalities. }
\vspace{0.5em}
\begin{tabular}{@{\;}ll@{\quad} ccc @{\quad} ccc@{\;}}
\toprule
 {Data Split} & & \multicolumn{4}{c}{ COCO 5-fold 1$\mathrm{K}$ Test~\cite{chen2015coco-cap}} \\
 {Eval Task} & & \multicolumn{2}{c}{\textsc{img}~$\rightarrow$~\textsc{text}} & \multicolumn{2}{c}{\textsc{text}~$\rightarrow$~\textsc{img}} \\
 Features & {Size Aug.} & R@1 & R@5 & R@1 & R@5 \\
\midrule
\multirow{3}{*}{\makecell[c]{\textsc{butd} (Region) \\ + BiGRU}} & $\varnothing$ & 74.6 & 95.0 & 60.6 & 89.1  \\
& visual & 75.4 & 95.5 & 61.4 & 89.5  \\
& visual+text & {78.5} & {96.0} & {61.7} & {90.3}  \\

\bottomrule
\end{tabular}
\label{tab:aug}
\end{table}

\mypara{(3) Different choices of the sequence model in GPO}
We also try using different sequence model to implement GPO. Table~\ref{tab:gpo_rnn} shows that using a Transformer Encoder~\cite{Vaswani2017Transformer} to replace the simple BiGRU does not yield improvements on two different combinations of features. The sequence model of GPO only takes the positional information as the input without using the exact feature vectors, thus we believe it's not necessary for the sequence model to have large capacity. A simple BiGRU will suffice for both capacity and computational efficiency, and more complex mechanisms like the multi-head attentions of Transformers could even hurt the performance.  
\begin{table}[t]
\centering
\tabcolsep 3pt
\small
\caption{Different choices of the sequence model used by GPO}
\vspace{0.5em}
\begin{tabular}{@{\;}lc@{\quad} ccc @{\quad} ccc@{\;}}
\toprule
 {Data Split} & & \multicolumn{4}{c}{ COCO 5-fold 1$\mathrm{K}$ Test~\cite{chen2015coco-cap}} \\
 {Eval Task} & & \multicolumn{2}{c}{\textsc{img}~$\rightarrow$~\textsc{text}} & \multicolumn{2}{c}{\textsc{text}~$\rightarrow$~\textsc{img}} \\
 Features & {Seq. Model} & R@1 & R@5 & R@1 & R@5 \\
\midrule
\multirow{2}{*}{\makecell[c]{\textsc{butd} (Region) \\ + BiGRU}} & Transformer & 77.4 & 95.4 & 61.4 & 90.1 &   \\
& BiGRU & {78.5} & {96.0} & {61.7} & {90.3}  \\
\midrule
\multirow{2}{*}{\makecell[c]{\textsc{butd} (Grid) \\ + BiGRU}} & Transformer & 76.6 & 95.7 & 62.7 & 90.7   \\
& BiGRU & {78.0} & {95.8} & {62.6} & {90.6}  \\
\bottomrule
\end{tabular}
\label{tab:gpo_rnn}
\end{table}

\subsection{More results and comparisons for image-text matching}
\begin{table*}[t]
    \centering
    \small
    \caption{Extension of Table~2 of the main text, with more image-text matching results on COCO and Flickr30K, using different visual and textual backbones (denoted by \textbf{bold section title}). $\star$: Ensemble results of two models; on IN/IN+VG: Models pre-trained on ImageNet~\cite{Russakovsky2014ImageNet}, ImageNet and VisualGenome~\cite{Krishna2016VisualGenome}, respectively. The best and second best results (in \symtext{rsum}) are marked \textbf{bold} in \first{red} and \second{black}.
    }
    {
        \tabcolsep 3.5pt
	    \begin{tabu}{@{\;} ll @{\quad\;\;} ccccccc ccccccc@{\;}}
	        \addlinespace
	        \toprule
	        {Data Split} & & \multicolumn{7}{c}{ COCO 5-fold 1$\mathrm{K}$ Test~\cite{chen2015coco-cap}} &
	        \multicolumn{7}{c}{Flickr30$\mathrm{K}$ 1$\mathrm{K}$ Test~\cite{young2014image}} \\ \cmidrule(lr){3-9} \cmidrule(lr){10-16}
	        {Eval Task} & &  \multicolumn{3}{c}{\textsc{img}~$\rightarrow$~\textsc{text}} & \multicolumn{3}{c}{\textsc{text}~$\rightarrow$~\textsc{img}} & & \multicolumn{3}{c}{\textsc{img}~$\rightarrow$~\textsc{text}} & \multicolumn{3}{c}{\textsc{text}~$\rightarrow$~\textsc{img}}  \\
	        {Method} & Feature Type & \symtext{r}{@1} & \symtext{r}{@5} & \symtext{r}{@10} & \symtext{r}{@1} & \symtext{r}{@5} & \symtext{r}{@10} & \symtext{rsum} & \symtext{r}{@1} & \symtext{r}{@5} & \symtext{r}{@10} & \symtext{r}{@1} & \symtext{r}{@5} & \symtext{r}{@10} & \symtext{rsum} \\
	        \midrule
	        \multicolumn{10}{@{\;}l}{\bf {ResNet-152 on IN~\cite{He2015ResNet} $+$ {BiGRU}}} \\[2pt]
	        UVS\cite{kiros2014UVS}\textsubscript{2014} & Grid & 56.0 & 85.8 & 93.5 & 43.7 & 79.4 & 89.7 & 448.1 & 42.1 & 73.2 & 84.0 & 31.8 & 62.6 & 74.1 & 367.8 \\
	        VSE++\cite{faghri2017vse++}\textsubscript{2017} & Grid & 64.6 & 90.0 & 95.7 & 52.0 & 84.3 & 92.0 & 478.6 & 52.9 & 80.5 & 87.2 & 39.6 & 70.1 & 79.5 & 409.8 \\
	        SCO\cite{huang2018SCO}\textsubscript{2018} & Grid & 69.9 & 92.9 & 97.5 & 56.7 & {87.5} & {94.8} & 499.3 & 55.5 & 82.0 & 89.3 & 41.1 & 70.5 & 80.1 & 418.5 \\
	        GXN$\star$\cite{gu2018GXN}\textsubscript{2018} & Grid & 68.5 & - & 97.9 & 56.6 & - & 94.5 & - & 56.8 & - & 89.6 & 41.5 & - & 80.1 & -  \\
	        Our: VSE++ & Grid & {70.9} & {93.4} & 97.5 & {58.2} & 87.1 & 93.5 & \second{500.6} & 64.4 & 87.3 & 93.1 & 49.3 & 77.5 & 84.7 & \second{456.3}  \\
	        Our: \ourmethod & Grid & {76.5} & {95.3} & {98.5} & {62.9} & {90.6} & {95.8} & \first{519.6} & {77.1} & {94.5} & {97.1} & {58.5} & {84.1} & {89.6} & \first{500.9} \\
	        \midrule
	        \multicolumn{10}{@{\;}l}{\bf {ResNet-101 Faster-RCNN on IN+VG (\textsc{butd})~\cite{anderson2017updown} $+$ {BiGRU}}} \\[2pt]
	        SCAN$\star$\cite{lee2018scan}\textsubscript{2018} & Region & 72.7 & 94.8 & 98.4 & 58.8 & 88.4 & 94.8 & 507.9  & 67.4 & 90.3 & 95.8 & 48.6 & 77.7 & 85.2 & 465.0  \\
	        LIWE~\cite{Wehrmann2019LanguageAgnosticVE}\textsubscript{2019} & Region & 73.2 & {95.5} & 98.2 & 57.9 & 88.3 & 94.5 & 507.6 & 69.6 & 90.3 & 95.6 & 51.2 & 80.4 & 87.2 & 474.3 \\
	        VSRN$\star$\cite{Li2019VSRN}\textsubscript{2019} & Region & 76.2 & 94.8 & 98.2 & {62.8} & 89.7 & 95.1 & 516.8 & 71.3 & 90.6 & 96.0 & 54.7 & 81.8 & 88.2 & 482.6 \\
	        CVSE~\cite{Wang2020CVSE}\textsubscript{2020} & Region & 69.2 & 93.3 & 97.5 & 55.7 & 86.9 & 93.8 & 496.4 & 70.5 & 88.0 & 92.7 & 54.7 & 82.2 & 88.6 & 476.7 \\
	        CAAN~\cite{Zhang2020ContextAwareAN}\textsubscript{2020} & Region & 75.5 & 95.4 & 98.5 & 61.3 & 89.7 & 95.2 & 515.6 & 70.1 & 91.6 & 97.2 & 52.8 & 79.0 & 87.9 & 478.6 \\
            IMRAM$\star$~\cite{Chen2020IMRAMIM}\textsubscript{2020} & Region & 76.7 & 95.6 & 98.5 & 61.7 & 89.1 & 95.0 & 516.6 & 74.1 & 93.0 & 96.6 & 53.9 & 79.4 & 87.2 & 484.2  \\
    	    Our: VSE++ & Region & 68.5 & 92.6 & 97.1 & 54.0 & 85.6 & 92.7 & 490.5 & 62.2 & 86.6 & 92.3 & 45.7 & 73.6 & 81.9 & 442.3 \\
    	    Our: \ourmethod & Region & {78.5} & {96.0} & {98.7} & {61.7} & {90.3} & {95.6} & 520.8 & {76.5} & {94.2} & {97.7} & {56.4} & {83.4} & {89.9} & 498.1 \\
    	    Our: \ourmethod & Grid &  78.0 & 95.8 & 98.5 & 62.6 & 90.6 & 96.0 & \second{521.5} & 77.9 & 93.7 & 97.4 & 57.5 & 83.4 & 90.2 & \second{500.2} \\ 
    	    Our: \ourmethod & Region+Grid & {80.0} & {97.0} & {99.0} & {64.8} & {91.6} & {96.5} & \first{528.8} & {80.7} & {96.4} & {98.3} & {60.8} & {86.3} & {92.3} & \first{514.8} \\
    	    \midrule
	        \multicolumn{10}{@{\;}l}{\bf {ResNet-101 Faster-RCNN on IN+VG (\textsc{butd})~\cite{anderson2017updown} $+$ {BERT}~\cite{Devlin2019BERT}}} \\[2pt]
    	    Our: VSE++ & Region & 67.9 & 91.9 & 97.0 & 54.0 & 85.6 & 92.5 & 488.9 & 63.4 & 87.2 & 92.7 & 45.6 & 76.4 & 84.4 & 449.7 \\
	        Our: \ourmethod & Region & {79.7} & {96.4} & {98.9} & {64.8} & {91.4} & {96.3} & {527.5} & {81.7} & {95.4} & {97.6} & {61.4} & {85.9} & {91.5} & {513.5} \\
	        Our: \ourmethod & Grid & {80.4} & {96.8} & {99.1} & {66.4} & {92.1} & {96.7} & \second{531.6} & {81.5} & {97.1} & {98.5} & {63.7} & {88.3} & {93.2} & \second{522.3} \\
	        Our: \ourmethod & Region+Grid & {82.2} & {97.5} & {99.5} & {68.1} & {92.9} & {97.2} & \first{537.4} & {85.3} & {97.2} & {98.9} & {66.7} & {89.9} & {94.0} & \first{532.0} \\
	        \midrule
	        \multicolumn{10}{@{\;}l}{\bf {ResNeXT-101 on IG (\textsc{wsl})~\cite{Mahajan2018WSL} $+$ {BERT}~\cite{Devlin2019BERT}}} \\[2pt]
	        Our: VSE++ & Grid & 79.6 & 97.1 & 99.0 & 66.4 & 91.1 & 95.5 & 528.7 & 80.9 & 96.6 & 98.9 & 65.2 & 89.5 & 93.7 & 524.8 \\
	        Our: \ourmethod & Grid & {84.5} & {98.1} & {99.4} & {72.0} & {93.9} & {97.5} & \second{545.4} & {88.4} & {98.3} & {99.5} & {74.2} & {93.7} & {96.8} & \second{550.9} \\
	        Our: \ourmethod$\star$ & Grid & {85.6} & {98.0} & {99.4} & {73.1} & {94.3} & {97.7} & \first{548.1} & {88.7} & {98.9} & {99.8} & {76.1} & {94.5} & {97.1} & \first{555.1} \\
    	    \bottomrule
	    \end{tabu}
	}
    \label{tab:extend_image_text}
\end{table*}
\begin{table*}[t]
    \centering
    \small
    \caption{Extension of Table~2 of the main text, with image-text matching results on COCO 5K. $\star$: Ensemble results of two models.
    }
    {
        \tabcolsep 5pt
	    \begin{tabu}{@{\;} ll @{\quad\;\;} ccccccc@{\;}}
	        \addlinespace
	        \toprule
	        {Data Split} & & \multicolumn{6}{c}{ COCO 5$\mathrm{K}$ Test~\cite{chen2015coco-cap}}  \\ 
	        {Eval Task} & & \multicolumn{3}{c}{\textsc{img}~$\rightarrow$~\textsc{text}} &  \multicolumn{3}{c}{\textsc{text}~$\rightarrow$~\textsc{img}} \\
	        {Method} & Feature Type & \symtext{r}{@1} & \symtext{r}{@5} & \symtext{r}{@10} & \symtext{r}{@1} & \symtext{r}{@5} & \symtext{r}{@10} & \symtext{rsum}  \\
	        \midrule
	        \multicolumn{4}{@{\;}l}{\bf {ResNet-152 on IN~\cite{He2015ResNet} $+$ {BiGRU}}} \\[2pt]
	        VSE++\cite{faghri2017vse++}\textsubscript{2017} & Grid & 41.3 & 71.1 & 81.2 & 30.3 & 59.4 & 72.4 & 355.7 \\
	        SCO\cite{huang2018SCO}\textsubscript{2018} & Grid & 42.8 & 72.3 & 83.0 & 33.1 & 62.9 & 75.5 & 369.6   \\
	        GXN$\star$\cite{gu2018GXN}\textsubscript{2018} & Grid & 42.0 & - & 84.7 & 31.7 & - & 74.6 & - \\
	        Our: VSE++ & Grid & 46.1 & 76.8 & 86.6 & 35.2 & 65.6 & 77.3 & \second{387.6}  \\
	        Our: \ourmethod & Grid & 55.1 & 81.9 & 89.9 & 40.9 & 70.6 & 81.5 & \first{419.9} \\
	        \midrule
	        \multicolumn{7}{@{\;}l}{\bf {ResNet-101 Faster-RCNN on IN+VG (\textsc{butd})~\cite{anderson2017updown} $+$ {BiGRU}}} \\[2pt]
	        SCAN$\star$\cite{lee2018scan}\textsubscript{2018} & Region & 50.4 & 82.2 & 90.0 & 38.6 & 69.3 & 80.4 & 410.9  \\
	        VSRN$\star$\cite{Li2019VSRN}\textsubscript{2019} & Region & 53.0 & 81.1 & 89.4 & 40.5 & 70.6 & 81.1 & 415.7 \\
	        CAAN~\cite{Zhang2020ContextAwareAN}\textsubscript{2020} & Region & 52.5 & 83.3 & 90.9 & 41.2 & 70.3 & 82.9 & 421.1  \\
            IMRAM$\star$~\cite{Chen2020IMRAMIM}\textsubscript{2020} & Region & 53.7 & 83.2 & 91.0 & 39.7 & 69.1 & 79.8 & 416.5  \\
    	    Our: VSE++ & Region & 42.9 & 74.5 & 85.1 & 31.7 & 61.8 & 74.2 & 370.2 \\
    	    Our: \ourmethod & Region & 56.6 & 83.6 & 91.4 & 39.3 & 69.9 & 81.1 & 421.9 \\
    	    Our: \ourmethod & Grid & 56.2 & 83.7 & 90.9 & 40.8 & 70.6 & 81.5 & \second{423.7}  \\ 
    	    Our: \ourmethod & Region+Grid & 59.8 & 86.1 & 92.8 & 42.7 & 72.8 & 83.3 & \first{437.5} \\
    	    \midrule
	        \multicolumn{7}{@{\;}l}{\bf {ResNet-101 Faster-RCNN on IN+VG (\textsc{butd})~\cite{anderson2017updown} $+$ {BERT}~\cite{Devlin2019BERT}}} \\[2pt]
    	    Our: VSE++ & Region & 42.1 & 72.6 & 83.9 & 31.0 & 61.3 & 73.7 & 364.7  \\
	        Our: \ourmethod & Region & 58.3 & 85.3 & 92.3 & 42.4 & 72.7 & 83.2 & {434.3} \\
	        Our: \ourmethod & Grid & 59.1 & 85.9 & 92.8 & 44.1 & 74.1 & 84.0 & \second{440.0} \\
	        Our: \ourmethod & Region+Grid & 62.5 & 87.8 & 94.0 & 46.0 & 75.8 & 85.7 & \first{451.8} \\
	        \midrule 
	        \multicolumn{7}{@{\;}l}{\bf {ResNeXT-101 on IG (\textsc{wsl})~\cite{Mahajan2018WSL} $+$ {BERT}~\cite{Devlin2019BERT}}} \\[2pt]
	        Our: VSE++ & Grid & 57.9 & 85.2 & 92.8 & 44.9 & 74.5 & 84.0 & 439.2 \\
	        Our: \ourmethod & Grid & 66.4 & 89.3 & 94.6 & 51.6 & 79.3 & 87.6 & \second{468.9} \\
	        Our: \ourmethod$\star$ & Grid & 68.1 & 90.2 & 95.2 & 52.7 & 80.2 & 88.3 & \first{474.8} \\
    	    \bottomrule
	    \end{tabu}
	}
    \label{tab:extend_coco_5k}
\end{table*}
In Table~\ref{tab:extend_image_text} and Table~\ref{tab:extend_coco_5k}, we provide results extending Table~2 of the main text. Grid feature with ImageNet-pretrained ResNet-152 was the standard image backbone for image-text matching before \textsc{butd} region features became popular. It is worth noting that our re-implementation of VSE++ improves the original VSE++ by a large margin, by increasing the image size from $224\times 224$ to $512\times 512$ as guided by the empirical study of~\cite{Jiang2020DefenseGridVQA}. \ourmethod consistently outperforms the improved VSE++ and other baselines on ResNet-152 grid features. 

We also include results of three non-VSE methods: SCAN~\cite{lee2018scan}, CAAN~\cite{Zhang2020ContextAwareAN}, and IMRAM~\cite{Chen2020IMRAMIM}. These methods rely on fine-grained cross-modality interactions to match image and text, and all of them use \textsc{butd} and BiGRU as the feature extractors.  Under the same experimental setups, \ourmethod outperforms them without any complex cross-modal modeling. 

\subsection{Synthetic Experiment for Verifying GPO Design}
We further verify the architecture of coefficient generator $g(\cdot,\cdot)$ using synthetic data and pre-determined pooling coefficients (\eg, coefficients for the K-MaxPool with $K=5$). As a concrete example, we generate a set of random feature vectors as the input to GPO, and then use the pre-determined "ground-truth" pooling strategy to generate the "ground-truth" output. Such synthetic input-output pairs are then used for learning a GPO module. As for evaluation, we took the coefficient generator from GPO and compare the predicted pooling coefficients to the "ground-truth" ones. We report the results in Root Mean Square Error (RMSE). 

\mypara{Evaluation Protocol} We designed four types of synthetic pooling patterns: (1). AvgPool (denoted as A), (2). K-MaxPool (denoted as M-K), (3). Top-K$\%$  Pooling (denoted as T-K$\%$), and (4). Linearly-decayed pooling weights (denoted as L), in which the pooling weight for the $k$-th maximum linearly decreases to zero as $k$ goes from $1$ to $\cN$. To better assess generalization, we set the training feature set sizes to range from 20 to 100, and we evaluate models on test data with the feature set size ranging from 10 to 120. We report results on both \textsc{Seen} and \textsc{Unseen} feature set sizes to investigate GPO's generalization performance.

\mypara{Different GPO Designs}
We compare different design choices of the architecture for $g(\cdot,\cdot)$, including:
\begin{itemize}[leftmargin=*,nolistsep]
	\item \textbf{Cos/Sin+BiGRU} This is the design introduced in the main paper, which uses the positional encoding and learn a BiGRU as the coefficients generator.
    \item \textbf{Interp.} Learn a fixed size vector as the pooling coefficients, and use linear interpolation to get the pooling coefficients for various lengths $\cN$. FSPool~\cite{Zhang2020FSPoolLS} uses this type of design to handle variable-length inputs. 
    \item \textbf{Cos/Sin+MLP} Instead of using a sequence model to handle variable-size features, simply using a MLP to map positional encodings into pooling coefficients. This design assumes the weight generation process for each index $k$ is unaware of the global length.
    \item \textbf{Index+BiGRU} This model transforms the position index into embeddings with a learnable matrix, and learns a BiGRU as the coefficients generator. Without the positional encoding, this design applies no prior knowledge to the ordinal position indices.
\end{itemize}

\begin{table*}[ht!]
\centering
\small
\caption{Comparisons of different GPO designs on synthetic patterns. Results are reported in RMSE (lower the better).}
\subfloat{
\begin{tabular}{llccccc}
\toprule
 \multirow{2}{*}{Input Repr.} & \multirow{2}{*}{Decoder}  & \multicolumn{5}{c}{\textsc{Seen Sizes}}  \\
  & & A & M-1 & M-10 & T-50$\%$ & L \\
\midrule
Index   & Interp. & \bf 0 & .065 & .030  & .011 & .004  \\
Cos/Sin   & MLP & .012 & .003 & .047 & .032 & .022 \\
Index   & BiGRU & .002 & \bf .002 & .026 & .011 & .008 \\
Cos/Sin   & BiGRU & \bf 0 &  .005 & \bf .010 & \bf .006 & \bf 0 \\
\bottomrule
\end{tabular}
}
\subfloat{
\begin{tabular}{ccccc}
\toprule
  \multicolumn{5}{c}{\textsc{Unseen Sizes} (smaller / larger)}  \\
  A & M-1 & M-10 & T-50$\%$ & L \\
\midrule
 \bf 0/0 & .017/.070 & .141/.014  & \textbf{.039}/.005 & .016/.002 \\
 .054/.006 & \textbf{.003}/.003 & .093/.030 & .096/.026 & .065/.028 \\
 .006/.003 &  .004/\textbf{.001} & .052/.018 & .049/.007 & .015/.005 \\
 .002/\textbf{0} & .010/.004 & \bf .031/.007 & .046/\textbf{.004} & \bf .005/.001 \\
\bottomrule
\end{tabular}
}
\label{tab:design}
\end{table*}

\mypara{Results}
Table~\ref{tab:design} presents the results comparing different design of GPOs. \emph{Cos/Sin+BiGRU} achieves the best overall performances. \emph{Interp.} has clear difficulty in handling K-Max Pooling. \emph{Index+BiGRU} produces slightly worse results, which shows the advantage of using \emph{Cos/Sin} positional encoding. Moreover, generalizing to unseen feature sizes is indeed challenging for GPO, and generalizing to smaller feature sizes is harder than generalizing to larger feature sizes. To make GPO better generalize to inputs with different sizes, we propose the Size Augmentation as discussed in \S~3 of the main paper. 

\subsection{Complex Variants of GPO}
We have kept a simple architecture for GPO so that it only adds marginal extra computational cost to VSE models. However, it is worthwhile to verify whether more complex variants of GPO can indeed produce better results.  We investigate two modifications: 

\mypara{Are per-dimension pooling coefficients helpful?} Instead of generating shared pooling coefficients for all dimensions, we try to generate coefficients for each dimension separately. Per-dimensional pooling has stronger capacity and might improve the performance, like how FSPool~\cite{Zhang2020FSPoolLS} is designed. However, results in Table~\ref{tab:gpo_per_dim} disproves the above statement in the context of \ourmethod. The per-dimension variant of GPO provides no improvements over two combinations of feature extractors, potentially due to over-fitting.
\begin{table}[t]
\centering
\small
\caption{Comparing variants of GPO w/ or w/o per-dimension coefficients.}
\vspace{0.5em}
\begin{tabular}{@{\;}lc@{\quad} cc @{\quad} cc@{\;}}
\toprule
 \multicolumn{2}{l}{Data Split}  & \multicolumn{4}{c}{ COCO 5-fold 1$\mathrm{K}$ Test~\cite{chen2015coco-cap}} \\
 \multicolumn{2}{l}{Eval Task} & \multicolumn{2}{c}{\textsc{img}~$\rightarrow$~\textsc{text}} & \multicolumn{2}{c}{\textsc{text}~$\rightarrow$~\textsc{img}} \\
 {Features} & Per-dim? & R@1 & R@5 & R@1 & R@5 \\
\midrule
\multirow{2}{*}{\makecell[c]{\textsc{butd} (Region) \\ + BiGRU}} & \cmark &  76.5 & 95.9 & 61.2 & 89.6 \\
 & \xmark & 78.5 & 96.0 & 61.7 & 90.3 \\
 \midrule
\multirow{2}{*}{\makecell[c]{\textsc{butd} (Region) \\ + BERT}} & \cmark & 79.3 & 96.1 & 64.7 & 91.3 \\
 & \xmark & 79.7 & 96.4 & 64.8 & 91.4 \\
\bottomrule
\end{tabular}
\label{tab:gpo_per_dim}
\end{table}

\mypara{Would GPO be better if $g(\cdot,\cdot)$ also takes feature as input?}  Another possible modification to GPO is to input both the feature itself and the position index into the coefficients generator $g(\cdot, \cdot)$. With this modification, GPO can be considered as a special form of self attention. By intuition, this modified GPO can adaptively change the pooling coefficients according to the exact feature values. However, Table~\ref{tab:gpo_data_dep} shows that this modification does not bring improvements. Position index along suffices for generating good pooling coefficients. 
\begin{table}[t]
\centering
\small
\caption{Comparing variants of GPO w/ or w/o data-dependent pooling coefficients generator.}
\vspace{0.5em}
\begin{tabular}{@{\;}lc@{\quad} cc @{\quad} cc@{\;}}
\toprule
 \multicolumn{2}{l}{Data Split}  & \multicolumn{4}{c}{ COCO 5-fold 1$\mathrm{K}$ Test~\cite{chen2015coco-cap}} \\
 \multicolumn{2}{l}{Eval Task} & \multicolumn{2}{c}{\textsc{img}~$\rightarrow$~\textsc{text}} & \multicolumn{2}{c}{\textsc{text}~$\rightarrow$~\textsc{img}} \\
 {Features} & \makecell{Feature to\\$g(\cdot,\cdot)$?} & R@1 & R@5 & R@1 & R@5 \\
\midrule
\multirow{2}{*}{\makecell[c]{\textsc{butd} (Region) \\ + BiGRU}} & \cmark & 78.0 & 96.2 & 61.8 & 90.2  \\
 & \xmark & 78.5 & 96.0 & 61.7 & 90.3 \\
 \midrule
\multirow{2}{*}{\makecell[c]{\textsc{butd} (Region) \\ + BERT}} & \cmark & 79.2 & 96.3 & 64.7 & 91.2 \\
 & \xmark & 79.7 & 96.4 & 64.8 & 91.4 \\
\bottomrule
\end{tabular}
\label{tab:gpo_data_dep}
\end{table}

In \S~3.1 of the main paper, we observe that complex aggregators cannot outperform well-selected simple pooling function,  and the above two experiments again show that complicated feature aggregation does not necessarily improve VSE models. A possible explanation for these experimental results is that: feature extractors have provided adequate information for multi-modal matching, so the feature aggregators do not have to further contextualize the feature vectors. Too complicated models for feature contextualization might increase the risk of over-fitting and hurt the performance at the end.

{\small
\bibliographystyle{ieee_fullname}
\bibliography{egbib}
}

\end{document}